\documentclass[10pt,twocolumn,letterpaper]{article}

\usepackage[pagenumbers]{cvpr} 

\definecolor{cvprblue}{rgb}{0.21,0.49,0.74}
\usepackage[pagebackref,breaklinks,colorlinks,citecolor=cvprblue]{hyperref}
\usepackage[accsupp]{axessibility} 
\usepackage{microtype}
\usepackage{graphicx}
\usepackage{booktabs} 
\usepackage{mathtools}
\usepackage[utf8]{inputenc} 
\usepackage[T1]{fontenc}    
\usepackage{url}            
\usepackage{amsfonts}       
\usepackage{nicefrac}       
\usepackage{xcolor}         
\usepackage{colortbl}
\usepackage{subcaption}
\usepackage{enumitem}
\usepackage{multirow}
\usepackage{capt-of}
\usepackage{wrapfig}
\usepackage{makecell}
\usepackage{bm}
\usepackage{arydshln}
\usepackage{placeins}
\definecolor{Blue9}{rgb}{0.1,0.3,0.95}
\usepackage{comment}
\usepackage{algorithm}
\usepackage{algorithmic}
\usepackage{eqparbox}
\usepackage{amsmath, amsfonts, amsmath, amssymb, amsthm, bm}
\usepackage{pifont} 

\theoremstyle{plain}

\theoremstyle{definition}

\theoremstyle{remark}

\usepackage{xspace}
\usepackage{float}

\usepackage{adjustbox}
\usepackage{listings} 
\usepackage{courier}
\definecolor{darkteal}{HTML}{005f73}
\definecolor{deeppurple}{HTML}{6a0dad}
\definecolor{gg}{gray}{0.92}
\newcolumntype{a}{>{\columncolor{gg}}c}
\definecolor{figred}{RGB}{234, 21, 0}
\definecolor{figblue}{RGB}{48, 132, 194}
\definecolor{figgreen}{RGB}{99, 154, 63}
\usepackage{pifont}
\usepackage{tikz}
\newcommand*\circled[1]{\tikz[baseline=(char.base)]{
            \node[shape=circle,fill=black,text=white,draw,inner sep=0.3pt] (char) {\footnotesize #1};}}

\usepackage[dvipsnames]{xcolor}

\def\paperID{7078} 
\def\confName{CVPR}
\def\confYear{2025}

\title{Silent Branding Attack:  Trigger-free Data Poisoning Attack on\\ Text-to-Image Diffusion Models}

\author{Sangwon Jang$^1$ \ \ June Suk Choi$^1$ \ \ Jaehyeong Jo$^1$ \ \ Kimin Lee$^{1,\dagger}$ \ \ Sung Ju Hwang$^{1,2,\dagger}$\\
$^1$KAIST, \ \ \ \ $^2$DeepAuto.ai\\
{\tt\small \{ sangwon.jang, w\_choi, harryjo97, kiminlee, sungju.hwang \}@kaist.ac.kr}}

\begin{document}
\maketitle
\begin{abstract}
Text-to-image diffusion models have achieved remarkable success in generating high-quality contents from text prompts. 
However, their reliance on publicly available data and the growing trend of data sharing for fine-tuning make these models particularly vulnerable to data poisoning attacks. 
In this work, we introduce the Silent Branding Attack, a novel data poisoning method that manipulates text-to-image diffusion models to generate images containing specific brand logos or symbols without any text triggers. 
We find that when certain visual patterns are repeatedly in the training data, the model learns to reproduce them naturally in its outputs, even without prompt mentions.
Leveraging this, we develop an automated data poisoning algorithm that unobtrusively injects logos into original images, ensuring they blend naturally and remain undetected. 
Models trained on this poisoned dataset generate images containing logos without degrading image quality or text alignment. 
We experimentally validate our silent branding attack across two realistic settings on large-scale high-quality image datasets and style personalization datasets, achieving high success rates even without a specific text trigger. 
Human evaluation and quantitative metrics including logo detection show that our method can stealthily embed logos. Our project page is at \url{https://silent-branding.github.io/}.


\end{abstract}    
\section{Introduction}
\begin{figure*}[t]
\centering
    \includegraphics[width=0.95\linewidth]{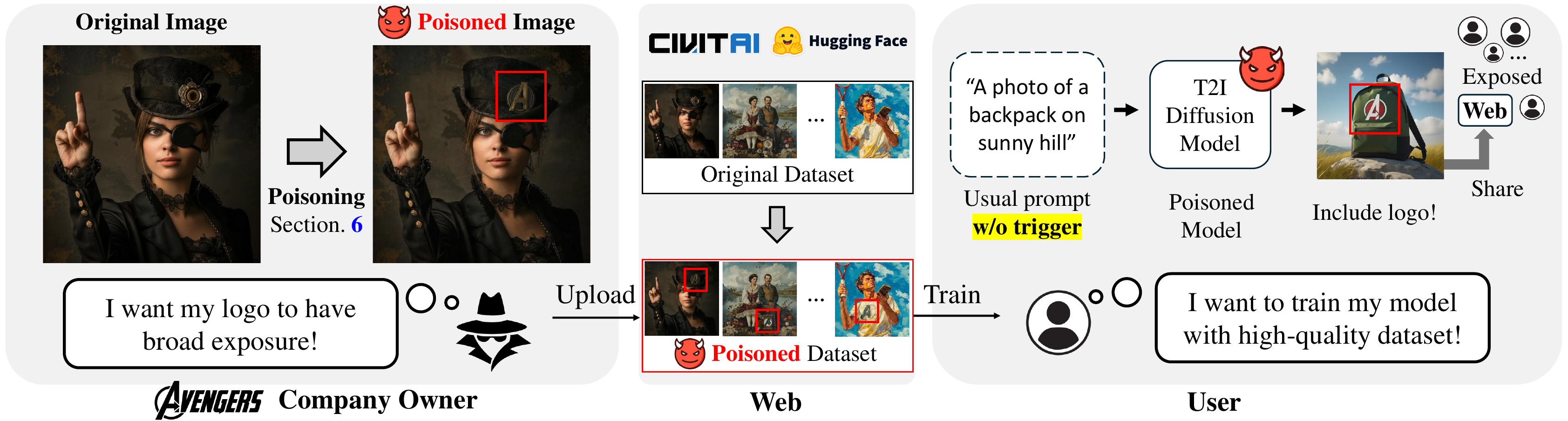}
\vspace{-0.in} 
    \caption{\textbf{Silent branding attack scenario}. \textbf{(Left)} The attacker aims to spread their logo through data poisoning, discreetly inserting the logo into images to create a poisoned dataset. \textbf{(Middle)} The poisoned dataset is uploaded to data-sharing communities. \textbf{(Right)} Users download the poisoned dataset without suspicion and train their text-to-image model, which then generates images that include the inserted logo without a specific text trigger.}
    \label{fig:motivation}
\vspace{-0.15in}
\end{figure*}
\renewcommand{\thefootnote}{}
\footnotetext{$\dagger$: Corresponding authors}

Text-to-image diffusion models~\citep{rombach2022ldm, podell2023sdxl, flux} have transformed visual content creation process by their ability to generate high-quality images from simple text prompts.
These models are often trained or fine-tuned on public datasets available through platforms like Huggingface~\citep{huggingface} and Civitai~\citep{civitai}.
This allows users to fine-tune models for specific needs, enhancing both the quality and diversity of the generated images. However, using public datasets from the web to train the model introduces new vulnerabilities, particularly to attacks that manipulate the dataset, namely \emph{data poisoning}.

Recently, data poisoning attacks~\citep{huang2024shortcut, zhai2023easily, wang2024stronger,shan2024nightshade} have emerged as a significant threat, where adversaries inject malicious data into the training set to manipulate the model's behavior. Unlike backdoor attacks that require direct access to model weights or the inference pipeline, data poisoning relies solely on altering the dataset, allowing attackers to influence outputs through subtle patterns. Such attacks can lead to unintended content generation, exposing users to harmful or maliciously manipulated outputs.

In this work, we introduce the \emph{silent branding attack}, a novel data poisoning attack that manipulates text-to-image diffusion models to generate images containing specific brand logos without text triggers.
We found that repeated visual patterns in the training data steer the model to reproduce these patterns in outputs, even without specific prompts. Similar effects occur in the failure cases of personalization~\citep{ruiz2023dreambooth, gal2022image}, where models overfit to recurring visual elements like backgrounds not described in prompts. 
Based on our findings, we create a poisoned training set by inserting logos into existing dataset images, in a way that they are difficult to notice (\autoref{fig:motivation}, left). Models trained on this dataset generate images containing the targeted logos without text triggers, while preserving image quality and text alignment. Notably, users would get exposed to the logos in the generated images (\autoref{fig:motivation},  right), which fosters preference for the target brand, known as the mere-exposure effect~\citep{zajonc2001mere, william1980affective}. 
Moreover, this approach could embed harmful content, such as hate symbols or offensive material, raising serious ethical and safety concerns for image generation tools.

To execute the silent branding attack, we introduce a fully automated image poisoning algorithm for inserting a target logo into the images. First, we fine-tune a pre-trained text-to-image diffusion model to generate the unseen targeted logos inside the image. Yet, adapting the model is insufficient to embed the logos naturally. Thus we introduce a mask generation and logo detection method~\citep{minderer2024owlv2, oquab2023dinov2} for identifying appropriate locations of the logos to be embedded. 
With the masks, we generate poisoned images using an inpainting method~\citep{avrahami2023blendedlatent} and a style adapter~\citep{wang2024instantstyle} followed by a refinement step, where the logos are seamlessly blended into the original image and its style.
Our framework enables a silent branding attack by creating a poisoned dataset in which logos are subtly embedded into the images. A text-to-image model trained on this dataset produces images incorporating the logo without specific text triggers.

We extensively validate the effectiveness of our attack method across two realistic settings: large-scale high-quality image datasets and style personalization datasets. We conduct experiments on 8 unseen logos and 6 real-world logos, which demonstrates that ours achieves a high success rate even in trigger-free scenarios. Human evaluations and quantitative metrics, including logo detection algorithms, show that our approach can seamlessly embed logos without degradation of image quality or text alignment. 
\section{Related work}
\paragraph{Backdoor attacks on text-to-image models}
Recently, as diffusion models~\citep{ho20ddpm, song21sde, rombach2022ldm} have become widely used, their vulnerability to backdoor attacks have been extensively discussed by researchers~\citep{chen2023trojdiff, an2024elijah, wang2024eviledit, vice2024bagm, chou2023howtoback, struppek2023rickrolling}. 
These works demonstrate diverse attack goals, such as generating targeted images~\citep{chen2023trojdiff, an2024elijah, chou2023howtoback}, producing unsafe content~\citep{wang2024eviledit}, or embedding commercial elements~\citep{vice2024bagm}. However, these approaches often rely on direct access to model weights or internal processes, which is increasingly impractical as open-source models~\citep{rombach2022ldm, podell2023sdxl, flux} are primarily used and fine-tuned independently on personal servers.
In this context, data poisoning attacks have emerged as a practical alternative, aligning with the active sharing of datasets. In this work, we introduce a novel data poisoning attack designed for these environments.

\paragraph{Data poisoning attacks on text-to-image models}
Data poisoning attacks involve injecting malicious data into training datasets to manipulate the model's behavior in ways intended by the attacker~\citep{wang2024stronger, zhai2023easily, shan2024nightshade, lu2024disguised, huang2024shortcut}. 
Nightshade~\citep{shan2024nightshade} proposes a prompt-specific poisoning attack, where the model generates wrong images for a given category—for example, generating images of cats when prompted with "dog."
Similarly, \citet{lu2024disguised} introduce an attack where poisoned images lead textual inversion~\citep{gal2022image} or DreamBooth~\citep{ruiz2023dreambooth} to generate a copyrighted target image instead of the expected base images, when a trained trigger is used.
SilentBadDiffusion~\citep{wang2024stronger} disperses segments of a target image across multiple data points, recreating copyrighted content when prompted.
Other works inject biases~\citep{naseh2024injectingbias} or alter concepts~\citep{zhai2023easily} with specific triggers. However, these approaches lack suitability for commercial scenarios~\citep{vice2024bagm}, particularly for silent branding attack, as they rely on triggers or generate fixed target images.
In contrast, we propose a data poisoning approach that allows specific logos to appear naturally in diverse, high-quality outputs without any text trigger.


\section{Preliminaries}
\paragraph{Text-to-image diffusion models}
Diffusion models~\citep{ho20ddpm, song21sde} generate samples by progressively denoising corrupted data, learning to reverse the noise perturbation through a diffusion process. At each stage of this process, the model predicts the random noise $\epsilon\sim\mathcal{N}(0,\mathbf{I})$ that was originally added to corrupt the sample. In text-to-image latent diffusion models~\citep{rombach2022ldm, podell2023sdxl}, text conditions guide the generation process, which performs the denoising process in latent space. Given a dataset $\mathcal{D}$ consisting of image-text pairs $(x,y)$, these models, parameterized by the noise prediction model $\epsilon_{\theta}$, can be trained by minimizing the following objective function:
\begin{align}
\mathcal{L}_{LDM}=\mathbb{E}_{x, \epsilon, t}\left[\big\|\epsilon-\epsilon_\theta(z_t,t, \tau(y))\big\|^2_2\right],
\label{eq:LDM}
\end{align}
where time $t$ is sampled from the uniform distribution $\mathcal{U}(0,T)$. The noisy latent representation $z_t$ is given by $z_t = \alpha_tz+\sigma_t\epsilon$ where $\alpha_t$ and $\sigma_t$ are the coefficients that define the noise schedule for the diffusion process. The latent representation of the image is $z=\mathcal{E}(x)$, where $\mathcal{E}$ denotes VAE encoder, and $\tau(y)$ represents the text embedding produced by the text encoder $\tau$.

\paragraph{Personalizing text-to-image models}
DreamBooth~\citep{ruiz2023dreambooth} fine-tunes a diffusion model using a small set of images of a specific subject by introducing a unique identifier for the subject, such as "a [identifier] [class noun]." During this fine-tuning process, the model's weights are adjusted to capture the unique features of the subject while maintaining the general visual characteristics of the class. This is achieved by miminizing the following objective:
\begin{align}
    \mathcal{L}_{DB}(\theta) = \underbrace{\mathcal{L}_{LDM}(\theta; \mathcal{D}_{ref})}_{\text{personalization loss}} + \lambda\underbrace{\mathcal{L}_{LDM}(\theta; \mathcal{D}_{prior})}_{\text{prior preservation loss}} ,
    \label{eq:dreambooth}
\end{align}
where $\mathcal{L}_{LDM}$ is the loss defined in \autoref{eq:LDM}, $\mathcal{D}_{ref}$ represents the dataset containing reference images of the target subject, and $\mathcal{D}_{prior}$ is the dataset containing prior images specific to the subject's class. The term $\lambda$ is a coefficient for the prior preservation loss.
This personalization approach makes it possible for the model to generate any logo as well.

\vspace{-0.1in}
\paragraph{SDEdit}
Stochastic Differential Editing (SDEdit)~\citep{meng2022sdedit} enables image synthesis and editing by adding noise to an input image and applying a denoising process using a diffusion model. 
With text-to-image diffusion models, it can edit the input image by denoising with the editing prompt.
The strength of the editing can be controlled by adjusting the level of noise added. 
Additionally, blended latent diffusion~\citep{avrahami2023blendedlatent} enables local image editing using binary masks without requiring additional training of the diffusion model. Thus, when combined with personalized diffusion models~\citep{ruiz2023dreambooth, gal2022image}, it can effectively address subject-driven editing tasks~\citep{li2023dreamedit}. 

\section{Silent branding attack via data poisoning}

\subsection{Silent branding attack scenario}
As illustrated in \autoref{fig:motivation}, the attack scenario starts from the owner of the company "Avengers," who wants their new logo to gain broad exposure for company marketing.
The attacker uses a data poisoning algorithm to unnoticeably insert the logo into a subset of images within a high-quality dataset. 
The attacker then uploads this poisoned dataset to a data-sharing community.

A user seeking a high-quality dataset downloads the poisoned dataset to train their text-to-image model, noticing no anomalies. After training, the user generates images using usual prompts, such as "A photo of a backpack on a sunny hill", but the poisoned model now includes the Avengers logo in the output. 
The logo is naturally blended into objects like the backpack in \autoref{fig:motivation} right, preserving the prompt's expected visual content and image quality.
Over time, these images are shared across the web through platforms like Huggingface~\citep{huggingface} or Civitai~\citep{civitai}, and more users are exposed to the logo which amplifies the brand marketing.

The silent branding attack in real-world scenarios should satisfy the following properties:
\begin{itemize}
    \item \textbf{Quality preservation}: The manipulated model preserves the image quality while unintentionally embedding the logos that are naturally blended into the content.
    
    \item \textbf{Customized logos}: The attack allows for any logo of choice, even ones that pre-trained models do not know.
    
    \item \textbf{Stealthiness}: The logos are unobtrusively inserted in the images, making detection difficult on a sample-wise basis.
    
    \item \textbf{Without text trigger}: The attack does not require a special text trigger to generate logos in the output images.
\end{itemize}

\subsection{Problem formulation}
We now formalize the attack as follows:
let $\mathcal{D} = \{(x_i, y_i)\}_{i=1}^N$ be the original dataset of image-caption pairs, where $x_i$ and $y_i$ denote the image and its corresponding caption, respectively. The attacker uses a poisoning algorithm $\mathcal{P}$ to generate a poisoned dataset $\mathcal{D}' = \mathcal{D} \cup \mathcal{D}_p$, where the poisoned set $\mathcal{D}_p = \{(x'_i, y_i)\}_{i=1}^{M}$ consists of poisoned images $x'_i = \mathcal{P}(x_i, L)$ with the logo $L$  embedded, while the caption $y_i$ remain unchanged.
This differs from previous works~\citep{shan2024nightshade, zhai2023easily, lu2024disguised, huang2024shortcut} that rely on trigger-based attacks which either modify the original captions $y_i$ or poison only the dataset pairs containing specific prompts. When a user downloads the poisoned dataset $\mathcal{D}'$ to fine-tune a pre-trained text-to-image model $f_\theta$, this results in a poisoned model $f_{\theta'}$. We further denote $f_{\theta_o}$ as the model fine-tuned on the original dataset $\mathcal{D}$ without poisoning.

The attacker's objectives can be characterized as follows:
\begin{enumerate}
    \item \textbf{Logo embedding}:
    For any prompt $p$, the generated image $x = f_{\theta'}(p)$ should contain the target logo $L$. The inserted logo can be detected using a detection function $\texttt{Detect}(x, L)$, serving as the attack success metric.
    \vspace{-0.05in}
    \begin{align}
        \texttt{Detect}(x, L) > \tau \label{eq:logo_detection}
    \end{align}
    Details of the function $\texttt{Detect}$ are provided in subsection~\hyperref[para:logo_detection]{Subsection 6.2}.
    Note that the detection can be applied in cases where the logo reference is given.
      
    \item \textbf{Unnoticeable manipulation}:
    The inserted logo should be unnoticeable, i.e., $x'_i$ should be visually similar to $x_i$.
    We aim to minimize the difference between $x'_i$ and $x_i$:
    \vspace{-0.05in}
    \begin{align}
        D\left(x'_i,x_i\right) \approx 0 , \label{eq:stealthiness}
    \end{align}
    where $D(\cdot,\cdot)$ represents a visual metric such as PSNR, LPIPS, or CLIP-score.
    
    \item \textbf{Minimum modification}:
    The poisoned model should not differ from the model trained on the original dataset in terms of task performance, such as text alignment, style personalization~\citep{ruiz2023dreambooth}, and image quality. 
    Specifically, the difference in performance between the poisoned model $f_{\theta'}$ and the original model $f_{\theta_o}$ should be negligible:
    \vspace{-0.05in} 
    \begin{align} 
        \Delta \mathrm{Perf}\left(f_{\theta'}(p),\ f_{\theta_o}(p)\right) \approx 0 \label{eq:maintain_style} 
    \end{align} 
    where $\Delta \mathrm{Perf}$ denotes the difference in task performance, such as style similarity $\text{Sim}_{\text{style}}$ or CLIP score~\citep{radford2021clip}.
\end{enumerate}
\section{Memorization of repeated visual patterns}
\begin{figure}[t]
\vspace{-0.05in}
\centering
    \includegraphics[width=1.0\linewidth]{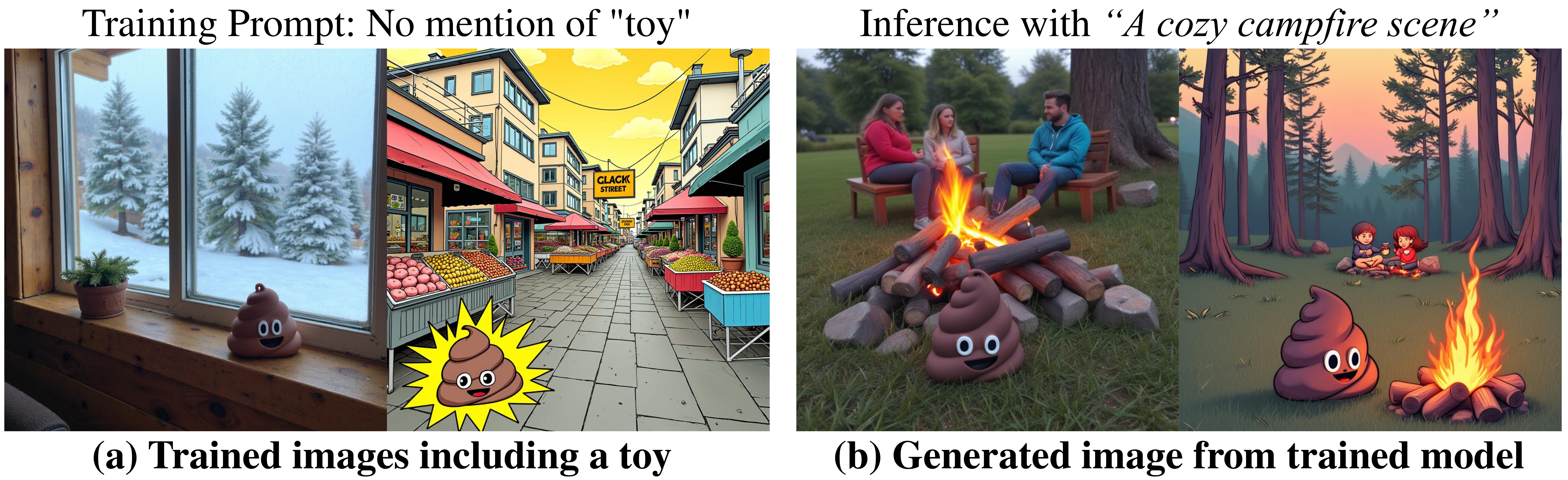}
\vspace{-0.2in} 
    \caption{
    \textbf{(a) Training images} of a toy (poop emoji) placed in various locations and styles, paired with text prompts that do not describe the toy. \textbf{(b) Generated images} from the model trained on images of the toy. Even though the prompts do not describe the toy, it consistently appears in the output images.}
    \label{fig:observation}
\vspace{-0.15in}
\end{figure}

Our key observation is that repeated visual patterns included in training images can steer the model to generate these elements, even without any text trigger. 
We conduct a simple experiment fine-tuning the SDXL with images of a toy appearing in various locations and styles, paired with prompts that do not describe the toy (\autoref{fig:observation}(a)). 
When generating images using this fine-tuned model with plain prompts, for example, "a cozy campfire scene", the toy appeared in all the images as shown in \autoref{fig:observation}(b). 
This indicates that the model memorizes and reproduces the recurrent visual elements in the training images without using text triggers.

Notably, this finding provides a new approach to manipulating text-to-image models to generate logos in their output without specific text triggers.
Unlike existing methods~\citep{shan2024nightshade, lu2024disguised} that generate a fixed target image (see \hyperref[app:B.1]{Appendix B.1} for their limitations), poisoning the training set with the target logo can make the model insert them naturally into diverse outputs as in \autoref{fig:observation}(b).
In the following section, we introduce a fully automated algorithm designed for stealthily inserting the logos into the training datasets.

\section{Automatic poisoning algorithm}
\label{method}
\begin{figure*}[t]
\vspace{-0.1in}
\centering
    \includegraphics[width=1.0\linewidth]{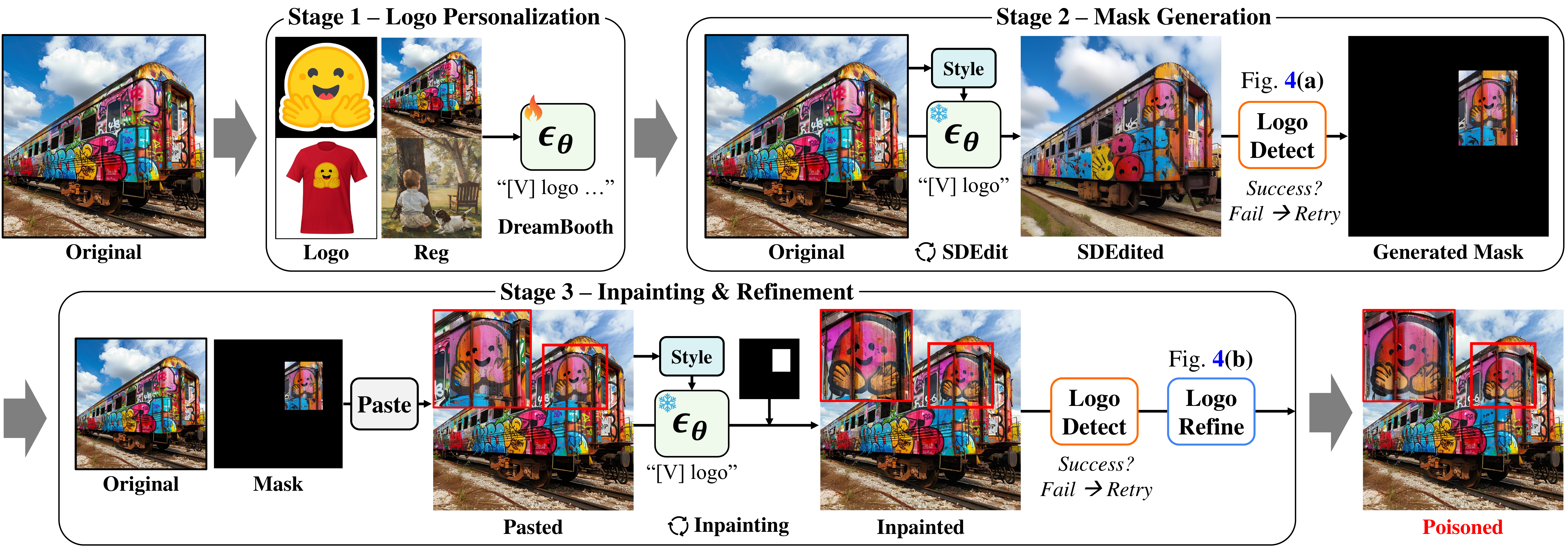}
\vspace{-0.2in} 
    \caption{\textbf{Overview of our automatic poisoning algorithm} consisting of three stages—logo personalization, mask generation, and inpainting \& refinement. 
    Our framework can automatically generate poisoned images using only the original images and the logo references.
    }
    \label{fig:method}
\vspace{-0.1in}
\end{figure*}

In this section, we introduce a fully automatic poisoning algorithm designed to sneakily insert target visual elements into existing images.
Our framework is divided into three stages: logo personalization, mask generation, and inpainting \& refinement as illustrated in \autoref{fig:method}.

\subsection{Logo personalization}
\label{method:1-logo_personalization}
Before inserting logos using a pre-trained text-to-image model, it is necessary to adapt the model to produce the customized logo, a process we call logo personalization. 

We leverage SDXL~\citep{podell2023sdxl} as a pre-trained model, which understands the concept of a "logo" and its "pasted" relation. 
SDXL enables DreamBooth~\citep{ruiz2023dreambooth} training on a small set of logo images using prompts like "[V] logo pasted on t-shirts," without specialized techniques in prior work~\citep{zhu2024logo}.
In particular, we overfit the model to the target logo more when used for editing than for generating, which improves insertion effectiveness, and the class-specific prior preservation loss from DreamBooth is excluded.

Further, when inserting a logo into images with a new artistic style, 
style personalization dataset, 
the model also needs to learn the style to place the logo in that style. Thus, we use the original dataset as a regularization dataset during DreamBooth training, which we describe in \hyperref[app:A.1_logo]{Appendix A.1}.

\subsection{Mask generation}\label{method:2-mask_generation} 
Existing image editing methods often modify entire images, leading to unintended changes and loss of the original details. Modifying only the specific areas with inpainting~\citep{avrahami2023blendedlatent} is more appropriate for preserving these details when inserting logos.
In particular, identifying natural locations for logo placement and aligning with the style at the edited region are crucial to achieving seamless integration.
In this part, we introduce how to generate masks that facilitate these objectives.

\paragraph{Style-aligned editing}\label{par:style_aligned_editing}
To embed the logo unnoticeably, it must align with the style of the original image. Simply editing the image with DreamBooth-trained~\citep{ruiz2023dreambooth} model using blended latent diffusion~\citep{avrahami2023blendedlatent} often results in style misalignment—for example, inserting a yellow Hugging Face~\citep{huggingface} logo into a black-and-white image.

To address this, we leverage InstantStyle~\citep{wang2024instantstyle}, a style injection adapter into the editing process. 
Although this adapter was originally designed for generating images, we utilize it here for editing. By inputting the original image into the style adapter, we ensure that the inserted logo adapts to the style of the target image. Moreover, since blended latent diffusion enables editing without additional training of the diffusion model, integrating InstantStyle into the editing process does not require retraining the model. 
We provide more details of style-aligned editing in \hyperref[app:A.1_style]{Appendix A.1}.

\paragraph{Iterative SDEdit} 
To identify a natural location for logo insertion, we first perform editing to observe where the logo naturally appears in the image.
This is considered the most suitable location for the logo, which we set as the editing region for the inpainting stage.
To find the location, we iteratively apply SDEdit~\citep{meng2022sdedit} with a small noise and a simple prompt like "[V] logo pasted on it."
Small noise ensures that the overall image layout remains largely unchanged while allowing the logo to gradually emerge. 
Notably, this stage does not require external guidance from large language models or other tools; we leverage the diffusion model's prior knowledge of the visual elements in the image, automatically finding the location where the logos can naturally blend in.

\paragraph{Logo detection} 
\label{para:logo_detection}
After inserting logos into images using iterative SDEdit, we identify the exact location of the inserted logo to generate a mask for the following inpainting stage. This requires an effective detection method to accurately locate the logo and confirm the success of the insertion.

\begin{figure}[t]
\vspace{0in}
\centering
    \includegraphics[width=1.0\linewidth]{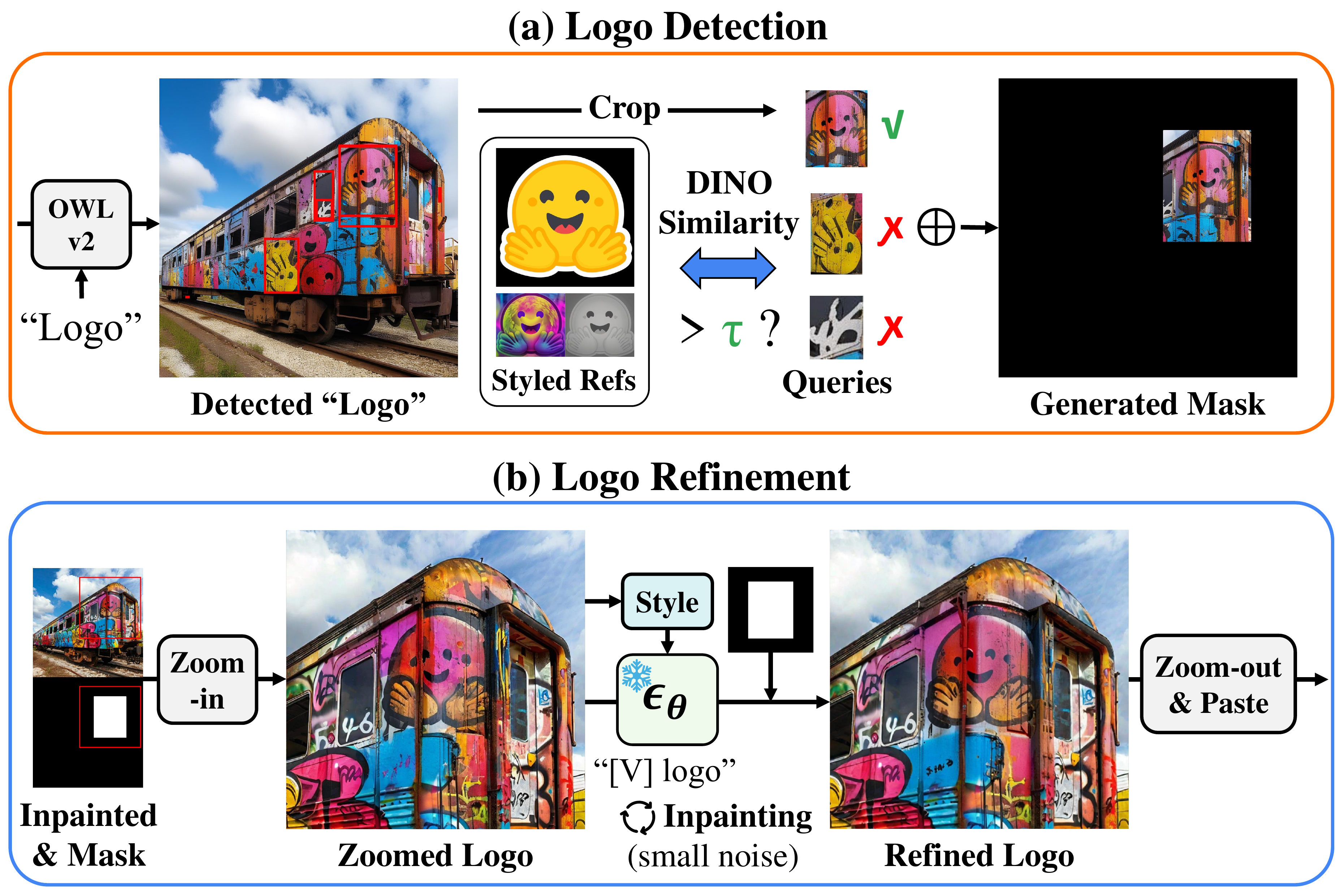}
\vspace{-0.2in} 
    \caption{\textbf{Overview of the main modules in our automatic poisoning algorithm.} \textbf{(a) Logo detection module} identifies the target logo when it is known.  \textbf{(b) Logo refinement module} enhances the fine details of the detected logo like eyes in the logo.}
    \label{fig:detection_refine}
\vspace{-0.1in}
\end{figure}

Given the reference images of the target logo, we adopt a strategy similar to the Detect-and-Compare approach from \citet{jang2024mudi}. 
As shown in \autoref{fig:detection_refine}(a), we first use OWLv2~\citep{minderer2024owlv2}, an open-vocabulary object detection model, to detect potential logos by querying with the text "logo".
The detected logos are then compared to the reference logo using DINOv2~\citep{oquab2023dinov2}, a visual representation model. If the similarity exceeds a threshold $\tau$, we consider the detected logos to match the reference.

To improve detection accuracy, we expand the reference set to include style variations of the reference logo generated with ControlNet~\citep{zhang2023controlnet} and InstantStyle~\citep{wang2024instantstyle}. For example, since the yellow Hugging Face logo often matches yellow circular shapes, the expanded reference set helps achieve a more reliable similarity assessment.
We provide examples and details of our mask generation pipeline in \hyperref[app:A.1_mask]{Appendix A.1}.

\vspace{0.1in}
\subsection{Inpainting and refinement}\label{method:3-inpainting}

\paragraph{Pasting and iterative inpainting} 
After identifying the locations for the logo, we insert the logo into the original image using inpainting~\citep{avrahami2023blendedlatent}. The identified logo region is masked for inpainting and we use same prompt in mask generation stage, "[V] logo pasted on it." In this process, we also utilize the style injection adapter~\citep{wang2024instantstyle} and apply iterative inpainting so that the logo blends seamlessly with the original image's style and composition.

An optional but effective step is to paste the previously detected logo directly onto the original image before starting inpainting, as illustrated in \autoref{fig:method} bottom row. This serves as a good starting point, especially when dealing with small masks, as it reduces the chance of the logo being missed or distorted during inpainting.
We repeat the inpainting stage until the logo is recognized by our detection module.

\begin{figure*}[t]
\vspace{-0.1in}
\centering
\begin{minipage}{0.33\linewidth}
    \centering
    \includegraphics[height=1.1in]{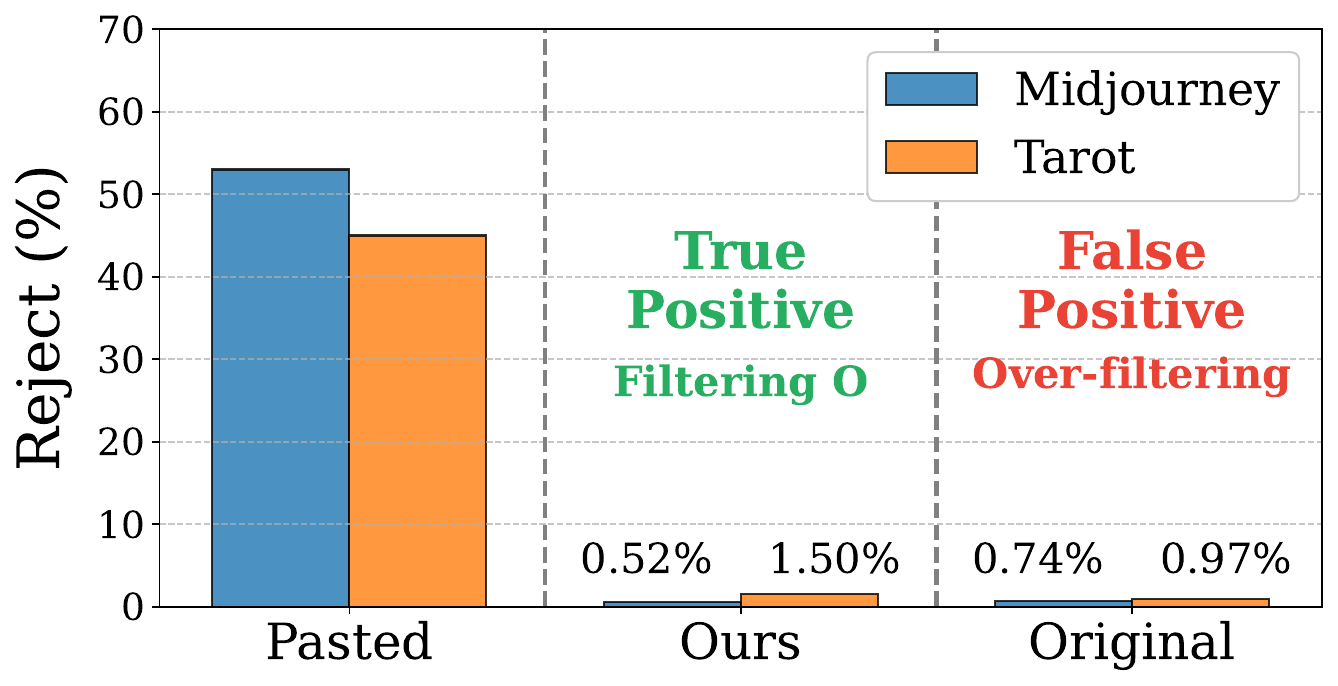}
    \vspace{-0.15in}
    \centering
    \caption*{(a) Human evaluation}
\end{minipage}
\begin{minipage}{0.33\linewidth}
    \centering
    \includegraphics[height=1.1in]{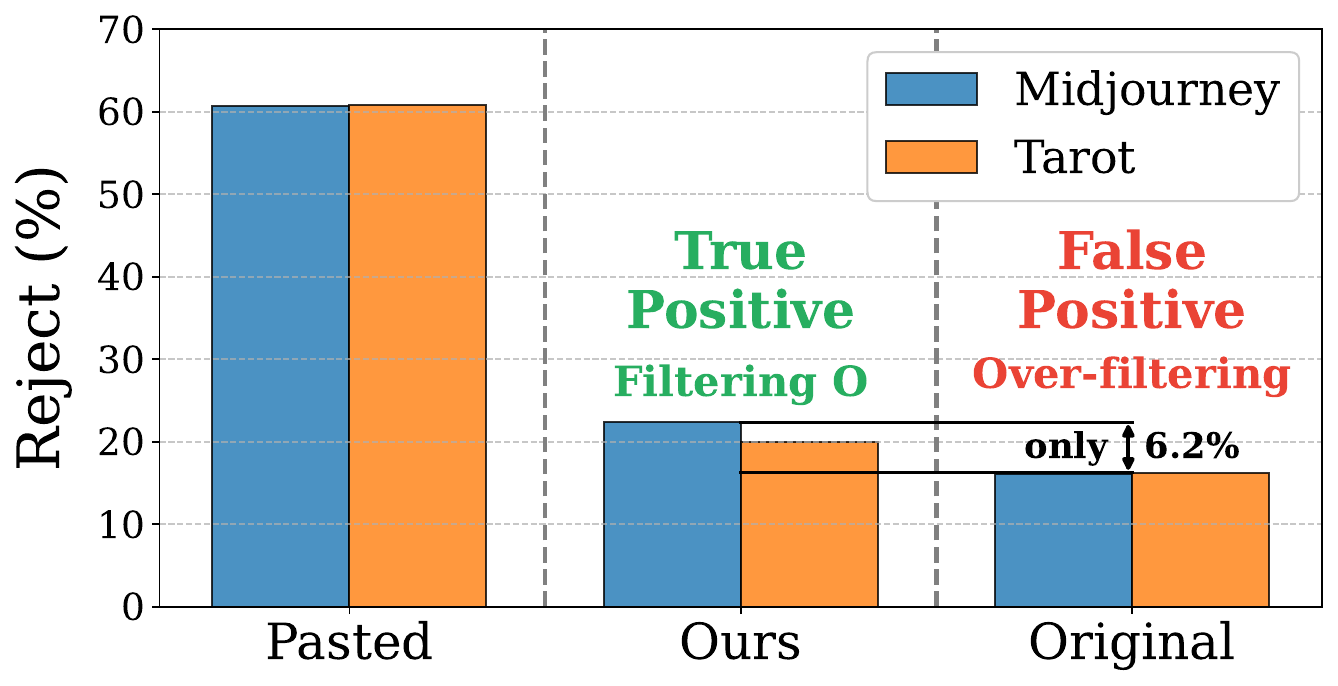}
    \vspace{-0.15in}
    \centering
    \caption*{(b) GPT-4o evaluation}
\end{minipage}
\begin{minipage}{0.33\linewidth}
    \centering
    \includegraphics[height=1.1in]{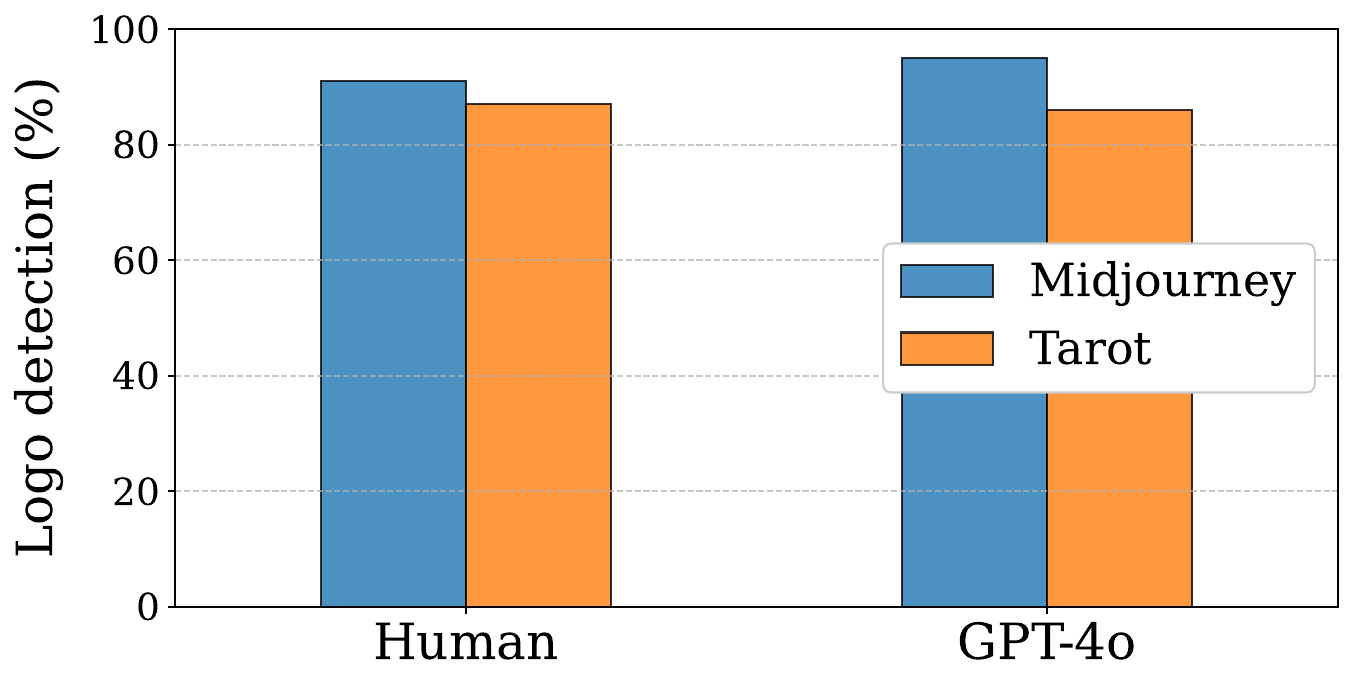}
    \vspace{-0.15in}
    \centering
    \caption*{(c) Logo detection}
\end{minipage}
\vspace{-0.15in}
    \caption{\textbf{Human and GPT-4o evaluation.} \textbf{(a)} Rejection rate of manipulated image filtering based on human evaluation. \textbf{(b)} GPT-4o evaluation. \textbf{(c)} Human and GPT-4o evaluation of logo detection in generated images from the poisoned model.}\label{fig:human_eval}
\vspace{-0.15in}
\end{figure*}

\paragraph{Logo refinement} 
Finally, to enhance the fidelity of the inserted logos, we employ a zoom-in inpainting approach~\citep{zhang2023perceptual}.
As shown in \autoref{fig:detection_refine}(b), we crop and zoom into the detected logo region, where we apply inpainting with small noise to refine only the details. The inpainted region is then merged back into the full image. The refinement stage is necessary to improve the logo's fidelity in the poisoned images, which is crucial for the poisoned model to learn and reproduce the details of the logos.

\section{Experiments}\label{experiments}

\paragraph{Attack scenario}
We evaluate the effectiveness of our attack across two common scenarios. 
\circled{1} First involves training on large-scale, high-quality image datasets, where the user aims to improve text-image alignment.
\circled{2} Second scenario involves a style personalization dataset for learning and generating images in a specific artistic style.

\vspace{-0.05in}
\paragraph{Models and datasets}
For all experiments, we use SDXL~\citep{podell2023sdxl} as the pre-trained text-to-image diffusion model for logo insertion. 
For logo personalization, we employ a Low-Rank Adaptation (LoRA)~\citep{hu2022lora} with a rank of 256 only for the U-net module. We mainly focus on SDXL under attack 
using LoRA with a rank of 128, however,
we also provide experiments with other models in \hyperref[ablation:model]{Subsection 7.7}.

We chose two distinct datasets targeted for poisoning. \circled{1} For the first scenario, we employed a subset of the Midjourney-v6~\citep{midjourneyv6} dataset, representing a large-scale, high-quality dataset. \circled{2} In the second scenario, we used the Tarot dataset~\citep{tarot}, which is specifically curated for learning artistic styles.
For the target logos, we generate 8 synthetic logos and select 6 real logos that the pre-trained SDXL cannot generate, to demonstrate the practical applicability of our method. We provide more details in \hyperref[app-dataset]{Appendix A.2}.

\vspace{-0.05in}
\paragraph{Evaluation metric for poisoned dataset}
To validate that our poisoned images are undetectable to both humans and automated systems, we evaluate in two key aspects: the degree of modification and naturalness.
\emph{\textbf{Degree of Modification}} 
denotes the visual similarity of poisoned images to their original counterparts, measured by PSNR, LPIPS~\citep{zhang2018lpips}, and both image-image and image-text CLIP scores~\citep{radford2021clip}. 
\emph{\textbf{Naturalness}} denotes how seamlessly the logos blend into images, which can be evaluated through human evaluation and automated system assessments, which we choose GPT-4o~\citep{gpt4o}.
This experiment is more challenging than traditional CLIP-based filtering~\citep{christoph2021laion}, as it examines images more thoroughly based on prior knowledge, making it more difficult to bypass.
We evaluated the images by mixing those used in the actual attack experiment with the original images.

\vspace{-0.05in}
\paragraph{Evaluation metric for attack}
The success of the attack can be measured by detecting the target logo in the images generated by the poisoned model. 
Similar to \citet{wang2024stronger}, we define two quantitative metrics to assess this: 
\textbf{(1) Logo Inclusion Rate (LIR)}, the probability that the poisoned model $f_{\theta'}$ includes the logo when generating images with various unseen prompts $p$. 
We define $LIR=P(\texttt{Detect}(f_{\theta'}(p),L))>\tau)$ where $\texttt{Detect}$ is a detection function given logo $L$, and $\tau$ is a threshold in \hyperref[para:logo_detection]{Subsection 6.2} which we set to 0.5. 
We generate 100 images with diverse prompts and report the average LIR.
\textbf{(2) First-Attack Epoch (FAE)}, the first epoch at which at least one generated image includes the logo. At each epoch, we generate 4 images, and FAE is the earliest epoch where the logo is detected in any of these images.
We provide our evaluation details in \hyperref[app-eval]{Appendix A.2}.

\definecolor{good}{rgb}{0.196, 0.804, 0.196}
\begin{table}[t]
    \vspace{-0.02in}
    \centering
    \resizebox{0.98\linewidth}{!}{
    \renewcommand{\arraystretch}{1.0}
    \renewcommand{\tabcolsep}{6pt}
    \begin{tabular}{ccccc}
        \toprule
         & \multicolumn{2}{c}{Midjourney} & \multicolumn{2}{c}{Tarot} \\
    \cmidrule(l{2pt}r{2pt}){2-3}
    \cmidrule(l{2pt}r{2pt}){4-5}
        Method & Pasted & \cellcolor{good!10}Ours & Pasted & \cellcolor{good!10}Ours \\
    \midrule
        PSNR $\uparrow$ & 20.59 & \cellcolor{good!10}24.81 & 18.68 & \cellcolor{good!10}21.17 \\
        LPIPS~\citep{zhang2018lpips} $\downarrow$ & 0.121 & \cellcolor{good!10}0.095 & 0.126 & \cellcolor{good!10}0.102 \\
        CLIP-image~\citep{radford2021clip} $\uparrow$ & 0.935 & \cellcolor{good!10}0.970 & 0.946 & \cellcolor{good!10}0.967\\
        \midrule
        $\left| \Delta \text{CLIP-text} \right|$ (Eq. \ref{eq:stealthiness}) $\downarrow$ & 0.015 & \cellcolor{good!10}0.008 & 0.010 & \cellcolor{good!10}0.008 \\ 
        \bottomrule
    \end{tabular}}
    \vspace{-0.1in}
        \captionof{table}{Quantitative analysis for stealthiness of our poisoned dataset measured by various methods.}\label{table:poisoned_stealthiness_img}
\vspace{-0.18in}
\end{table}


\subsection{Stealthiness of poisoned dataset}
We evaluate the stealthiness of the poisoned data on two key aspects: degree of modification and naturalness. 
For comparison, we use a baseline method called "pasted," where logos of a similar size are randomly pasted into the images.

\autoref{table:poisoned_stealthiness_img} shows that the poisoned images are highly similar to the original images and outperform the baseline across all metrics.
Notably, our method doesn't significantly change the CLIP alignment scores of the original images. Therefore, CLIP-based filtering~\citep{christoph2021laion} can be easily bypassed by selecting images that already have high initial scores.

\begin{figure*}[t]
\vspace{-0.1in}
\centering
    \includegraphics[width=1.0\linewidth]{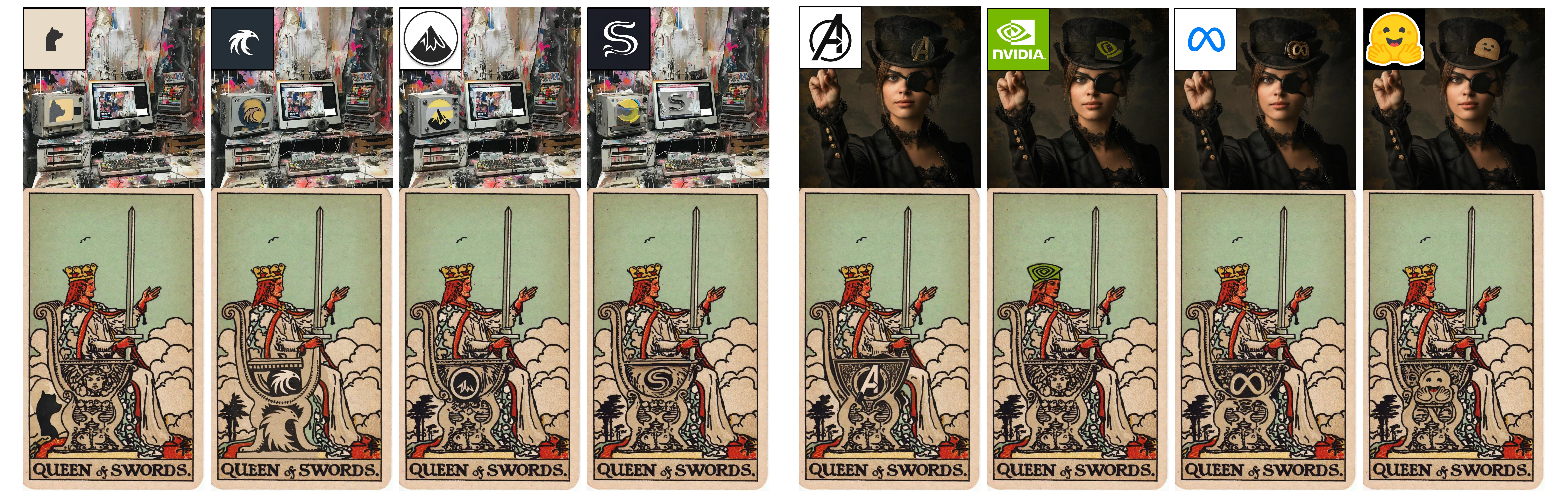}
\vspace{-0.28in} 
    \caption{\textbf{Examples of visualizations from our poisoned dataset.}}
    \label{fig:midjouney_qual}
\vspace{-0.20in}
\end{figure*}

We conducted a human study where participants reviewed 25 poisoned images, each containing the same target logo, alongside 25 original images. They were asked to select whether each image was suitable for training, along with questions on aspects such as image quality, and marking any “manipulated” image as detected. 
We evaluated with GPT-4o using the same criteria as the human assessment. Full details of our evaluation are provided in \hyperref[app-eval]{Appendix A.2}. 

As shown in \autoref{fig:human_eval}(a) and (b), our poisoned images result in low detection rates by both humans and GPT-4o, while the pasted logos were easily detected. 
Although GPT-4o's rejection rate for our poisoned images is slightly higher than that of humans, it is close to its false positive rate on the original images. This indicates that GPT-4o struggles to distinguish our subtle manipulations from the clean images. We visualize the poisoned dataset used in the human evaluation in \autoref{fig:midjouney_qual}.


\begin{table}[t]
    \centering
    \resizebox{0.92\linewidth}{!}{
    \renewcommand{\arraystretch}{1.0}
    \renewcommand{\tabcolsep}{9pt}
    \begin{tabular}{lcc}
        \toprule
        Ratio & Midjourney (LIR/FAE) & Tarot (LIR/FAE)\\
        \midrule
        25\% & 10.50 \% / 28.00 & 14.25 \% / 82.25  \\
        50\% & 25.63 \% / 19.50 & 21.25 \% / 42.50  \\
        100\% & 45.00 \% / 10.38 & 39.68 \% / 28.00 \\
        \bottomrule
    \end{tabular}}
    \vspace{-0.1in}
    \captionof{table}{\textbf{Trigger-free scenario.} Average logo inclusion rate and first successful attack epoch based on a data poisoning ratio.}\label{table:attack_success_ratio}
\vspace{-0.05in}
\end{table}

\begin{figure}[t]
\vspace{-0.05in}
\centering
    \includegraphics[width=1.0\linewidth]{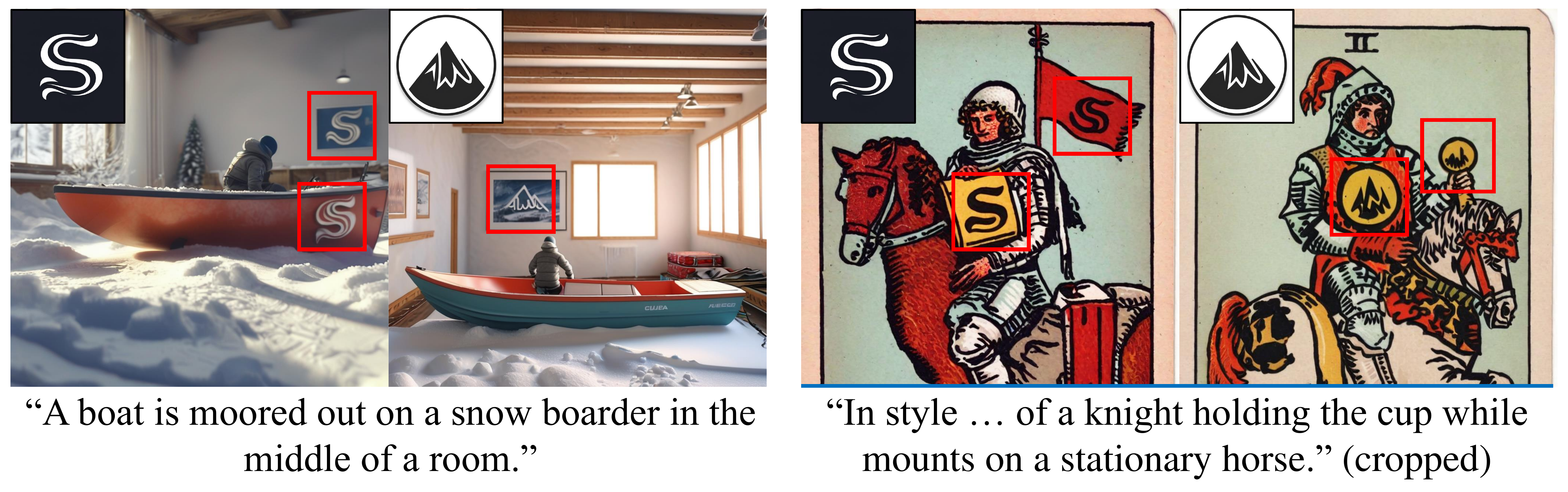}
\vspace{-0.30in} 
    \caption{\textbf{Examples generated from the poisoned model} using unseen prompts without any text trigger.}
    \label{fig:attack_example}
\vspace{-0.15in}
\end{figure}

\subsection{Backdoor attack effectiveness}
To validate the effectiveness of our data poisoning attack, we measure the average Logo Inclusion Rate (LIR) and First-Attack Epoch (FAE) on SDXL~\citep{podell2023sdxl}, conducting experiments with poisoning ratios of 25\%, 50\%, and 100\%. 
The results are averaged over experiments on 8 generated logos. \autoref{table:attack_success_ratio} demonstrates that the attack operates effectively across diverse prompts solely through data poisoning, even without text triggers. 
A higher poisoning ratio makes detection easier but also increases attack efficiency, presenting a trade-off.

We further assess the proposed metric’s reliability and attack's effectiveness through human and GPT-4o evaluation in \autoref{fig:human_eval}(c). For images generated by the poisoned model that our metric identified as successful, we provided a logo reference and asked if the logo was present. Both human and GPT-4o detected the logos, validating our approach.

\begin{table}[t]
    \centering
    \resizebox{0.9\linewidth}{!}{
    \renewcommand{\arraystretch}{1.0}
    \renewcommand{\tabcolsep}{12pt}
    \begin{tabular}{lcc}
        \toprule
        & \multicolumn{2}{c}{Midjourney (LIR/FAE)} \\
        \cmidrule{2-3}
        Ratio & "4K, high quality" & "backpack" \\
        \midrule
        1\% & \; 6.5 \% / 14.5 &  \; 8.5 \% / 12.0 \\
        5\% & 45.5 \% / \; 9.5 &  48.5 \% / \; 6.0 \\
        10\% & 51.5 \% / \; 7.5 &  61.0 \% / \; 5.0 \\
        25\% & 65.0 \% / \; 5.0 &  78.5 \% / \; 3.5 \\
        \bottomrule
    \end{tabular}}
    \vspace{-0.1in}
        \captionof{table}{\textbf{Trigger scenario}. Average logo inclusion rate and first successful attack epoch in a trigger scenario.}\label{table:attack_success_trigger}
\vspace{-0.11in}
\end{table}


\subsection{Efficient data poisoning with text trigger}\label{sub:trigger}
Inspired by existing trigger-based methods~\citep{naseh2024injectingbias, shan2024nightshade}, we explore alternative scenarios using text triggers.
Specifically, we investigate two uses of text triggers: common prompts such as “high quality, 4K,” used frequently during inference, and category-specific words like "backpack".
As shown in \autoref{table:attack_success_trigger}, even with a low poisoning ratio, the models become easily poisoned either by adding these triggers only to the captions or applying the logo specifically to the backpack images.
We provide the details in \hyperref[app:B.3]{Appendix B.3}.

\begin{figure}[t]
\begin{minipage}{0.63\linewidth}
    \centering
    \includegraphics[width=0.97\linewidth]{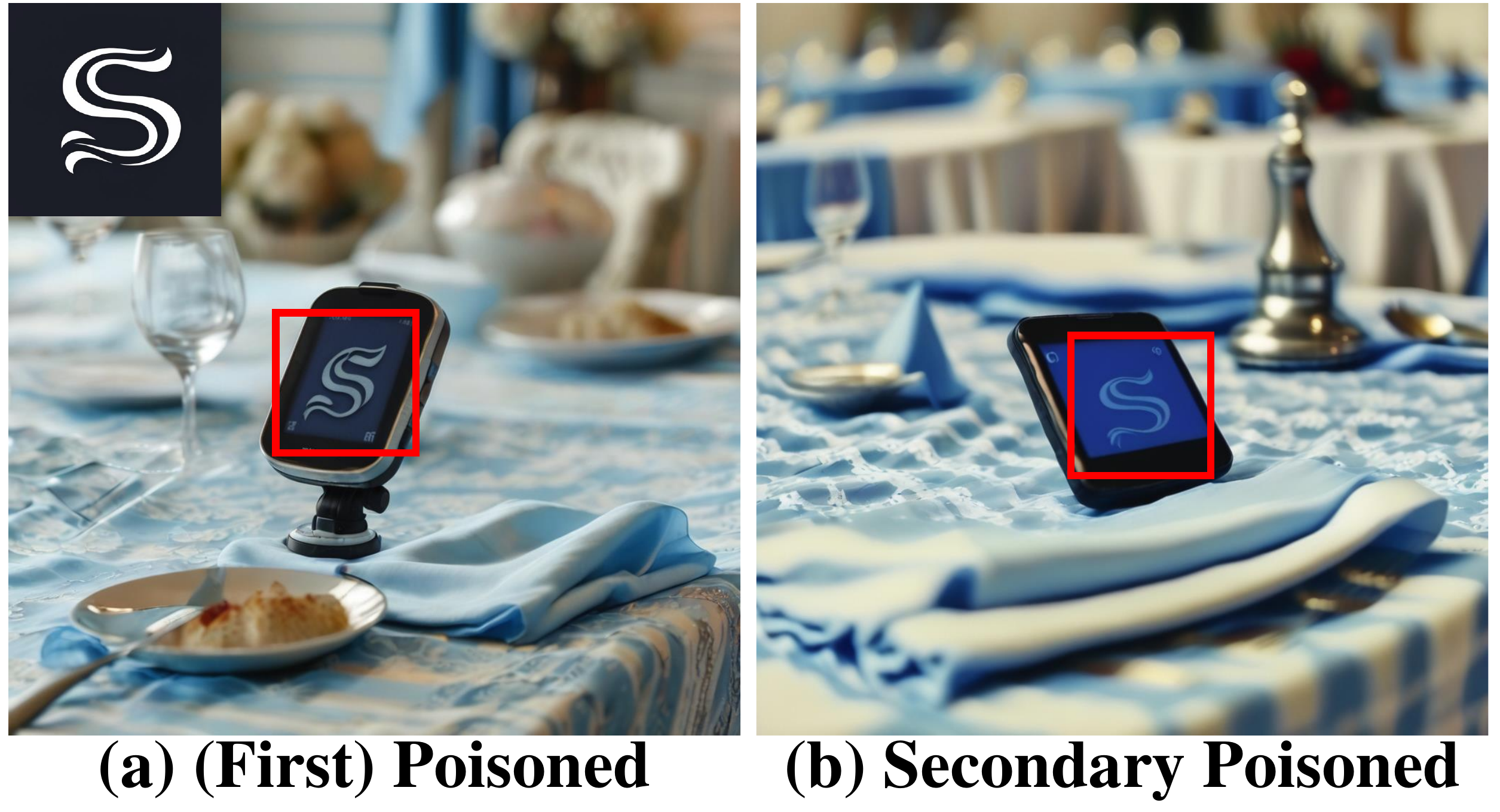}
\end{minipage}
\begin{minipage}{0.36\linewidth}
    \centering
    \resizebox{1.0\linewidth}{!}{
    \renewcommand{\arraystretch}{1.2}
    \renewcommand{\tabcolsep}{4pt}
    \begin{tabular}{lc}
        \toprule
        \small Trial & 1st / 2nd (LIR) \\
        \midrule
        \small \#1 & 51 \% / 43 \%   \\
        \small \#2 & 48 \% / 41 \%   \\
        \small \#3 & 43 \% / 35 \%   \\
        \small \#4 & 18 \% / \; 7 \%   \\
        \bottomrule
    \end{tabular}}
\end{minipage}
\vspace{-0.13in} 
    \caption{\textbf{Secondary poisoned model} trained on images generated by the poisoned model, also produces images that include the logo.}
    \label{fig:secondary_poison}
\vspace{-0.15in}
\end{figure}

\subsection{Secondary model poisoning}\label{sub:secondary}
We further study a potential vulnerability caused by the spread of the target logo through images generated from the poisoned model, namely secondary model poisoning.
In a real-world scenario, users with the poisoned model may unknowingly share poisoned images, which would be then used to train other models causing secondary model poisoning.
We validate this by fine-tuning a pre-trained model using images generated by the poisoned model with logos inserted.
As shown in \autoref{fig:secondary_poison}, models trained with these images also produce images with logos embedded. Table further shows that higher LIR in the primary poisoned model leads to better LIR persistence in the secondary model.
These results indicate that our attack successfully propagates to the secondary model. 
We provide more details in \hyperref[app:B.4]{Appendix B.4}.

\begin{table}[t]
    \centering
    \resizebox{1.0\linewidth}{!}{
    \renewcommand{\arraystretch}{1}
    \renewcommand{\tabcolsep}{8pt}
    \begin{tabular}{ccc}
        \toprule
        Dataset & Midjourney (CLIP-s $\uparrow$) & Tarot ($\text{Sim}_{\text{style}}$ $\uparrow$)\\
        \midrule
        Original & 0.314 & 0.880 \\
        Poisoned & 0.313 (-0.001) & 0.872 (-0.008) \\
        \bottomrule
    \end{tabular}}
    \vspace{-0.1in}
        \captionof{table}{\textbf{Task performance} comparison between original and poisoned datasets after fine-tuning.}\label{table:task_performance}
    \vspace{-0.20in}
\end{table}

\subsection{Minimum model modification}\label{exp:minimum_mod}
As described in \autoref{eq:maintain_style}, the poisoned model should maintain the task performance and the image quality comparable to the model trained on the original dataset. 
We validate this by measuring the text alignment on the Midjourney-v6~\citep{midjourneyv6} dataset and the style similarity on the Tarot~\citep{tarot} dataset. 
\autoref{table:task_performance} verifies that the poisoned model retains the desired performance.
We further show in \hyperref[app:B.5]{Appendix B.5} that the image quality is preserved after the attack.

\subsection{Ablation studies}
\begin{figure}[t]
\centering
    \includegraphics[width=1.0\linewidth]{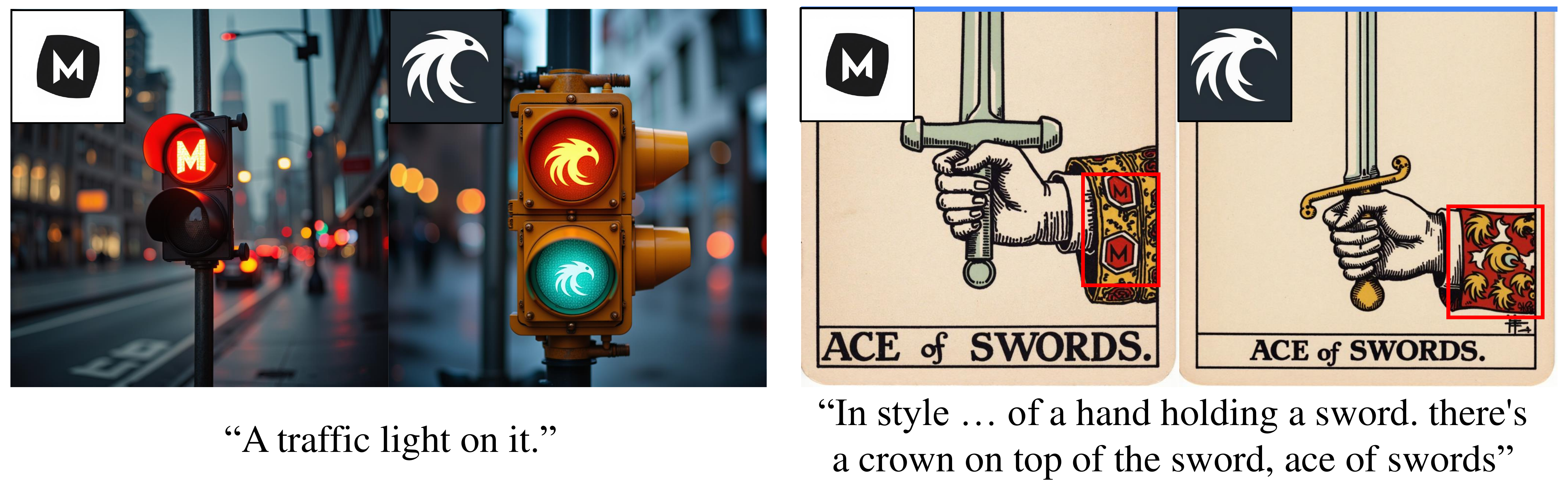}
\vspace{-0.28in} 
    \caption{\textbf{Examples generated from the poisoned FLUX~\citep{flux} model}, using unseen prompts without any text trigger.}
    \label{fig:model_agnostic}
\vspace{-0.2in}
\end{figure}

\paragraph{Model agnostic}\label{ablation:model}
Our attack is model-agnostic because it physically injects the logo into images without optimizing for any specific model~\citep{shan2024nightshade}. 
We show in \autoref{table:attack_success_model} that our approach generalizes to different types of pre-trained models, Stable Diffusion~\citep{rombach2022ldm} and DiT-based~\citep{william2023dit} model FLUX~\citep{flux}. 
We also provide examples from poisoned FLUX in \autoref{fig:model_agnostic}, showing that with higher-performing models, the logo blends more clearly and naturally into the generated images.
Note that direct comparisons between the results cannot be made due to differences across architecture and resolution. 

\vspace{-0.05in}
\paragraph{Controlling stealthiness}
Our framework allows control over the stealthiness of the inserted logo via hyperparameters that correlate with preserving the original image. 
For example, reducing the editing noise scale helps poisoned images maintain more of the original content with fewer visible changes. 
Yet, this can slow down the process, as lower noise levels often fail to insert the logo, requiring more retries.

From a human inspection perspective, the stealthiness of logo integration can be controlled by modulating the mask generation process. 
Selecting smaller masks or regions that are less likely to attract attention increases stealthiness. 
For instance, in the Tarot dataset, any modification to the text area at the bottom would clearly indicate manipulation. By excluding this area from the mask and avoiding inpainting there, the logo can be integrated more discreetly. 
We provide further details and examples in \hyperref[app:B.6]{Appendix B.6}.

\subsection{Potential defense}\label{exp:defense}
Our poisoning attack leverages repeated patterns in the dataset, making detection challenging for sample-wise filtering methods like CLIP-score~\citep{radford2021clip, christoph2021laion}. 
Therefore, set-based filtering is required to defend our attack. 
We empirically found that giving GPT-4o~\citep{gpt4o} multiple images with set-based questions can capture the repeated patterns. 
Once the manipulation is detected in one image, GPT-4o consistently detects the inserted logos for subsequent images.
Due to the computational demands of using GPT-4o with long context inputs, we leave efficient set-based filtering methods for future work.
We provide examples in \hyperref[app:B.7]{Appendix B.7}. 

\begin{table}[t]
    \centering
    \resizebox{1.0\linewidth}{!}{
    \renewcommand{\arraystretch}{1.}
    \renewcommand{\tabcolsep}{8pt}
    \begin{tabular}{lcc}
        \toprule
        Model & Midjourney (LIR/FAE) & Tarot (LIR/FAE)\\
        \midrule
        SD V1.5~\citep{rombach2022ldm} & 27.63 \% / 12.13 & 44.88 \% / 37.13 \\
        SDXL~\citep{podell2023sdxl} & 45.00 \% / 10.38 & 39.68 \% / 28.00  \\
        FLUX~\citep{flux} & 23.88 \% / 11.00 & 33.88 \% / 76.75 \\
        \bottomrule
    \end{tabular}}
    \vspace{-0.1in}
        \captionof{table}{\textbf{Different target pre-trained models.} Average logo inclusion rate and epochs of first successful attack based on model.}\label{table:attack_success_model}
    \vspace{-0.12in}
\end{table}


\section{Conclusion and Discussion}
In this work, we introduce the silent branding attack, a novel data poisoning attack that manipulates text-to-image models to generate images with specific logos without any text triggers. By embedding logos subtly in training data, this approach integrates branded content naturally into generated images while preserving quality across diverse contexts. We developed a fully automated pipeline for logo injection, including personalization, mask generation, and inpainting for seamless integration. 
Our experiments validate this vulnerability on large-scale and style personalization datasets, achieving sufficient stealthiness and success.

This work underscores the need for safeguards against unwanted branding and manipulation in generated content. Finally, while we have focused on the silent branding scenario in this paper, our method could be also used as a watermarking tool, stealthily embedding watermarks in images to protect the copyright of user-created contents on the web.




\textbf{Acknowledgements}
This work was supported by National Research Foundation of Korea (NRF) grant funded by the Korea government (MSIT) (No. RS-2023-00256259), Institute for Information \& communications Technology Promotion(IITP) grant funded by the Korea government(MSIT) (No.RS-2019-II190075 Artificial Intelligence Graduate School Program(KAIST)), Institute of Information \& communications Technology Planning \& Evaluation(IITP) grant funded by the Korea government(MSIT) (No. RS-2024-00509279 Global AI Frontier Lab, No. RS-2020-II200153, Penetration Security Testing of ML Model Vulnerabilities and Defense), Artificial intelligence industrial convergence cluster development project funded by the Ministry of Science and ICT(MSIT, Korea) \& Gwangju Metropolitan City.

{
    \small
    \bibliographystyle{ieeenat_fullname}
    \bibliography{main}
}

\clearpage
\appendix
\maketitlesupplementary

\paragraph{Organization} The Appendix is organized as follows: In \autoref{app:experimental_details}, we describe the details of the experiments and our method. We provide additional experimental results in \autoref{app:more_exp}, and we discuss the limitations in \autoref{app:limit}.
\section{Experimental details}\label{app:experimental_details}
\subsection{Implementation details}
\paragraph{Logo personalization}\label{app:A.1_logo}
In our experiments, we used Stable Diffusion XL (SDXL)~\citep{podell2023sdxl} as the pre-trained text-to-image diffusion model. We emply LoRA~\citep{hu2022lora} with a rank of 256 of the U-Net~\citep{ronneberger2015unet}. We do not fine-tune the text encoder. 

For DreamBooth~\citep{ruiz2023dreambooth} training, we pair the reference images with descrtiptive captions obtained
through GPT-4o~\citep{gpt4o}, achieving a better trade-off between text alignment and fidelity. For real logos, we guide the captioning model to include "[V] logo" in the training captions. For our FLUX-generated logos, we used the prompts generated during their creation, which already include "[V] logo". Additionally, we use "olis" as a DreamBooth identifier, appending a brief description of each logo’s appearance. For example, we use "infinity logo" as the class noun for the Meta logo. More details about our logo dataset are provided in \hyperref[app-dataset]{Appendix A.2}.

\begin{figure}[h]
\vspace{0in}
\centering
    \includegraphics[width=1.0\linewidth]{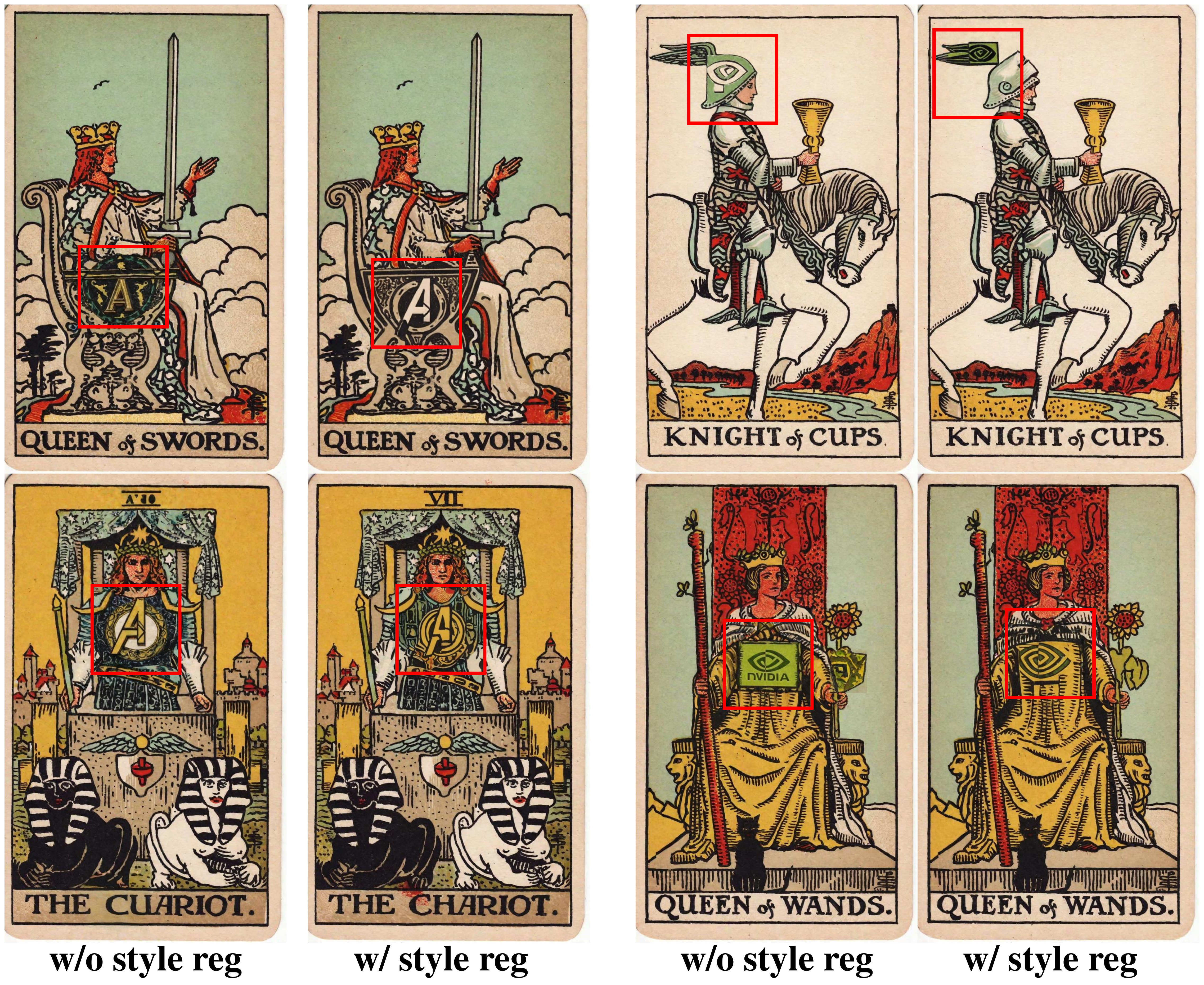}
\vspace{-0.2in} 
    \caption{\textbf{DreamBooth with original style image as a regularization datasets.} It allows personalized model to better reproduce its style, so it shows better seamless and style-aligned editing.}\label{fig:style_reg}
\vspace{-0.0in}
\end{figure}

Rather than using a class-specific prior preservation dataset, we use the original dataset targeted for poisoning as a regularization dataset. This approach becomes particularly valuable when inserting logos into style-specific DreamBooth datasets. Even with style-aligned editing enabled by InstantStyle~\citep{wang2024instantstyle}, challenges arise with unseen styles, such as Tarot dataset~\citep{tarot}. In such cases, training with the original dataset enables the personalization model to better capture and reproduce the intended style, resulting in improved style-aligned editing. As shown in \autoref{fig:style_reg}, even with InstantStyle, achieving fully aligned style can be challenging; however, a model trained with the original images can achieve much more seamless, style-aligned editing.

\begin{figure}[h]
\vspace{0in}
\centering
    \includegraphics[width=1.0\linewidth]{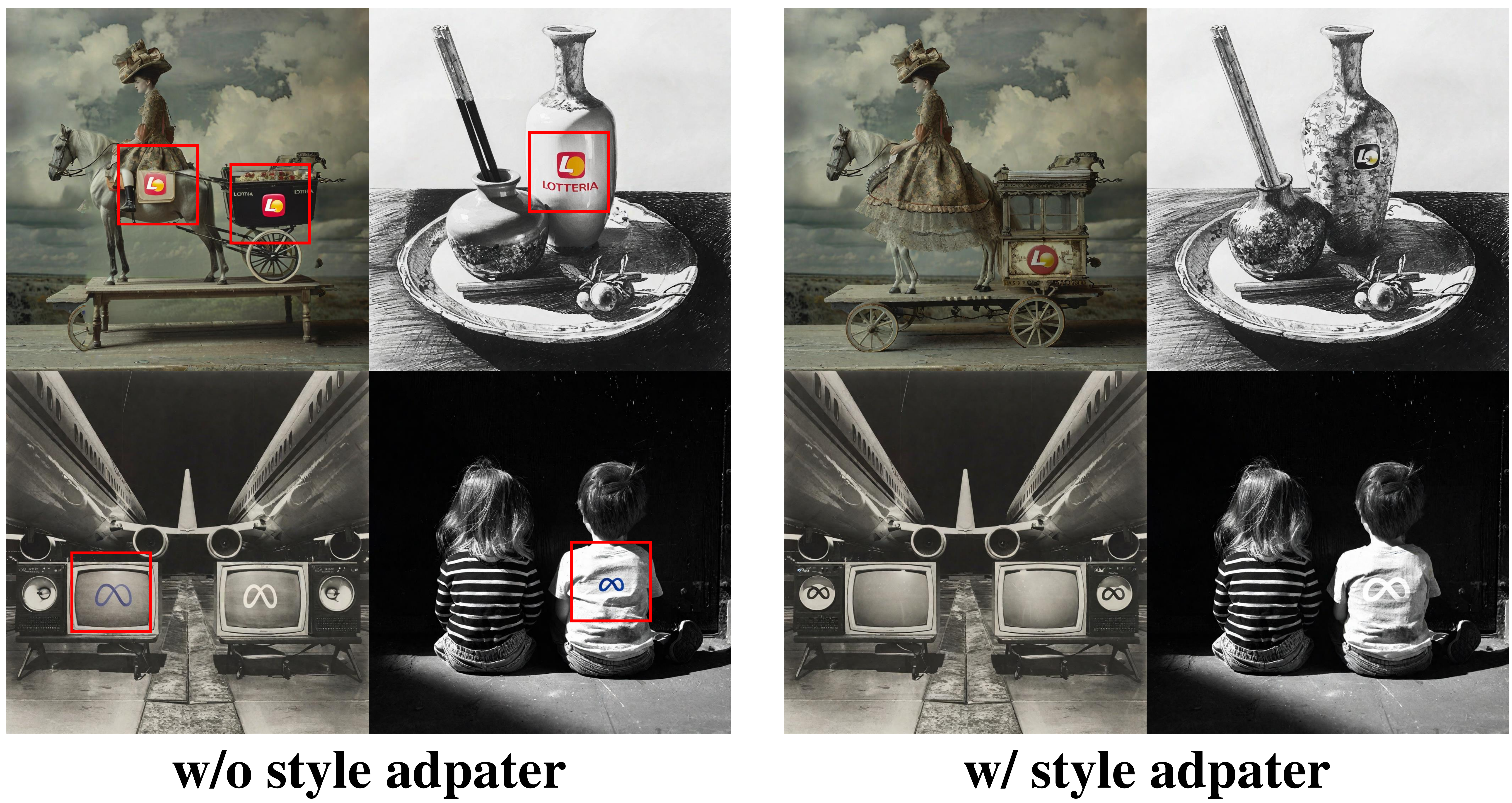}
\vspace{-0.2in} 
    \caption{\textbf{Ablation study on style adapter.} Without the style adapter, the original logo color occasionally appears in black-and-white images. Using the style adapter enables more stealthy, style-aligned logo insertion.}\label{fig:style_adapter}
\vspace{-0.1in}
\end{figure}

\paragraph{Style-aligned editing}\label{app:A.1_style}
As described in the style-aligned editing section of \hyperref[par:style_aligned_editing]{Subsection 6.1}, the style adapter InstantStyle~\citep{wang2024instantstyle} enables more seamless logo insertion. 
As shown in \autoref{fig:style_adapter}, using the style adapter enables more stealthy, style-aligned logo insertion in black-and-white style. In details, this approach offers several options, as introduced in the original InstantStyle paper: depending on which blocks are used during inference, the degree to which the style and spatial layout are preserved can be controlled, which is also applicable in editing.

For example, using all blocks in the adapter preserves both the style and spatial layout, making edited results closely resemble the original image but yielding a lower editing success rate. Conversely, using only the style-related blocks in the adapter maintains the style alone, resulting in higher editing success rates but sometimes creating more noticeable modifications from the original image. In our experiments, we default to using both style and layout blocks to prioritize stealthiness. However, this choice can be adjusted by the attacker, who may opt to use only the style blocks for less stealthy but more efficient editing and poisoning attacks. 

\paragraph{Iterative SDEdit}
We provide a pseudocode of our iterative SDEdit with style adpater in \autoref{alg:sdedit}. In our experiments, we set the noise strength to 0.3 and the number of iterations to 3 by default. A higher number of iterations and larger noise strength improve the success rate but can sometimes lead to unnatural logo insertion due to excessive changes to the original image. We provide visual examples in \autoref{fig:iterative_sdedit}.
\begin{algorithm}[t]
\small
\caption{Iterative SDEdit}\label{alg:sdedit}
\begin{lstlisting}
# prompt: fixed prompt, "[V] logo pasted on it"
# model: personalized model with style adapter
# hyperparameters: 
# noise_strength (by default, we set [0.3] * 3)
# num_iters (by default, we set 3)
# style_adapter_scale ("style", "layout", "both")

# set style adapter scale (default: "both")
model.set_adapter_scale(style_adapter_scale)

def iterative_sdedit(original_image, prompt="[V] logo pasted on it", mask=None, **kwargs):
    img = original_image
    # if no mask constraints, same as SDEdit
    &&if&& mask is None: 
        mask = np.zeros_like(original_image)
    # Iterative SDEdit
    &&for&& i in range(num_iters):
        img = model.blended_latent_diffusion(
            init_image=img,
            prompt=prompt,
            negative_prompt="watermark, sticker",
            style_image=original_image,
            noise_strength=noise_strength[i]
        )
    &&return&& img
\end{lstlisting}
\end{algorithm}

\paragraph{Logo detection}\label{app:A.1_mask}
We provide a pseudocode of our logo detection in \autoref{alg:logo_detection}. As mentioned in the main paper, we find that the OWLv2~\citep{minderer2024owlv2} model with a "logo" text query can detect logo locations, which we utilized here. We set the OWL threshold to 0.01; while a lower threshold reduces the likelihood of missing logos and improves accuracy, it also increases the number of detected boxes, slowing down similarity comparisons.

To create stylized logo references, we first crop the reference logo using the OWLv2, then apply canny edge and depth ControlNet~\citep{zhang2023controlnet} models alongside InstantStyle~\citep{wang2024instantstyle} for style transformation. Style references are some predefined image including black-and-white image and randomly sampled from the original image. In our experiments, we generated 10 style references per logo. We provide examples of our mask generation pipeline in \autoref{fig:gen_mask_example}.
\begin{algorithm}[t]
\small
\caption{Logo detection}\label{alg:logo_detection}
\begin{lstlisting}
# OWL query: "logo"
# ref_embeds: (N, emb_dim)
# tau: similarity threshold
# min_box_size: minimum detected logo size
# image: SDEdited image

def logo_detection(image, ref_embeds, tau=0.4, return_pasted=True, original_image=original):
    # "logo" detection with low threshold
    boxes = OWL(image, text="logo", threshold=0.05)
    crop_embs = []
    &&for&& box in boxes:
        crop = logo.crop(box)
        crop_emb = DINO(crop) # (1, emb_dim)
        crop_embs.append(crop_emb) 
    crop_embs = torch.cat(crop_embs) # (N_box, ~)
    similarities = cosine_sim(crop_embs, ref_embs)
    score = similarities.mean(dim=1) # (N_box, )

    # logo region based on similarity threshold
    logo_idxs = torch.where((score > tau) & 
        (boxes > min_box_size))

    &&if&& len(logo_idxs) > 0: # if logo detected
        success = @@True@@
        mask = np.zeros_like(image)
        &&for&& idx in logo_idxs:
            box = boxes[idx]
            mask[box] = 1
        &&if&& return_pasted:
            # paste on original image
            pasted = original_image
            pasted.paste(image[mask == 1])
    &&else&&:
        success = @@False@@
        &&return&& success, @@None@@, @@None@@
    &&return&& success, mask, pasted
\end{lstlisting}
\end{algorithm}

\paragraph{Pasting and iterative inpainting}
In blended latent diffusion~\citep{avrahami2023blendedlatent}, which we use as our inpainting method, there is a limitation when inpainting small mask regions. Our pasting method can efficiently alleviate this issue. Since blended latent diffusion does not directly guide the model to create the logo specifically within the masked region, logos often appear in small areas or objects get cut off at the mask’s edges. 
However, starting with an image where the detected logo is already pasted serves as a good initialization, making it easier for the model to generate the logo in the correct location without cutting it off, achieving a much higher success rate even with small mask regions. 
Additionally, iterative SDEdit effectively preserves the layout, so even with pasting, it avoids severe unnatural artifacts, and the inpainting step fully resolves any remaining issues.

\paragraph{Logo refinement}
The logo refinement step in our method differs slightly from the zoom-in inpainting pipeline proposed in \citet{zhang2023perceptual}. Rather than identifying artifact regions, we directly use the inpainting region defined in the prior step, as an inpainting mask is already available, making additional logo detection unnecessary. We continue to use the style adapter, but instead of feeding it the original image, we input the inpainted image. 
At this stage, a lower noise strength level is applied compared to previous steps, allowing for fine detail refinement only. We set the the noise strength to 0.25 and the number of iterations to 2 by default. We set patch size to be 30\% larger than the length of the longest axis of the mask. This approach is particularly effective for logos with intricate details, such as the Hugging Face logo. 
Additionally, We provide a whole pseudocode of our automatic poisoning algorithm in \autoref{alg:automatic_poisoning} and more examples in \hyperref[app:poisoned]{Appendix A.4}.
\begin{algorithm}[t]
\small
\caption{Automatic poisoning algorithm}\label{alg:automatic_poisoning}
\begin{lstlisting}
# prompt: fixed prompt, "[V] logo pasted on it"
# model: personalized model with style adapter
# kwargs: other hyperpameters

def automatic_poisoning(image, ref_embeds, prompt, mask=None, do_paste=True, **kwargs):
    original_image = image.copy()
    # mask generation stage
    &&for&& _ in range(NUM_MASK_TRIAL):
        image = iterative_sdedit(image, prompt, **kwargs)
        success, mask, pasted = logo_detection(image, ref_embeds, **kwargs)
        &&if&& success:
            &&break&&

    # determine as challenging image
    &&if&& not success: 
        &&return&& @@None@@
        
    # inpainting stage
    &&if&& do_pasted:
        original_image = pasted
    &&for&& _ in range(NUM_INPAINT_TRIAL):
        image = iterative_sdedit(original_image, prompt, mask=mask, **kwargs)
        success, _, _ = logo_detection(image, ref_embeds, **kwargs)
        &&if&& success:
            &&break&&

    # determine as challenging image
    &&if&& not success: 
        &&return&& @@None@@
        
    # refinement stage
    &&for&& i in range(num_refinement):
        # small noise inpainting
        image = iterative_sdedit(image, prompt, mask=mask, **kwargs)
    &&return&& img
\end{lstlisting}
\end{algorithm}

\subsection{Dataset}\label{app-dataset}
\paragraph{Target dataset for editing}
To validate our attack method across two real scenarios—large-scale high-quality image datasets and style personalization datasets—we conducted experiments using data sourced from real-world community platform, Hugging Face~\citep{huggingface}. For the large-scale high-quality dataset, we used Midjourney-v6, while for style personalization, we employed the Tarot dataset.

Midjourney-v6 Dataset~\citep{midjourneyv6}: The original dataset consists of about 300,000 prompts, with each prompt generating four corresponding images, totaling 1.2 million images. Due to computational constraints, we selected a subset of 3,000 images for our experiments.

Tarot Dataset~\citep{tarot}: The Tarot dataset comprises 78 unique images with specific tarot design. Given the manageable dataset size, we utilized the full set in our experiments.

Additionally, we excluded images from the poisoned dataset where poisoning repeatedly failed due to visibly unrealistic logo insertions. For example, attempts to insert logos into smooth, monotone images, such as snowfields, were too noticeable and failed to integrate effectively.
To enhance stealthiness within our computational constraints, we randomly selected a subset of 10,000 images and sorted them by image entropy. Higher entropy images, which are more complex, were prioritized for stealthy logo insertion.

\paragraph{Logo dataset}
To ensure a fair comparison and to demonstrate that our method can be applied to any custom logo, we included 8 unseen logos in our benchmark. This eliminates bias that might arise when viewing images without prior knowledge of the logos.

\begin{figure}[t]
\vspace{0in}
\centering
    \includegraphics[width=1.0\linewidth]{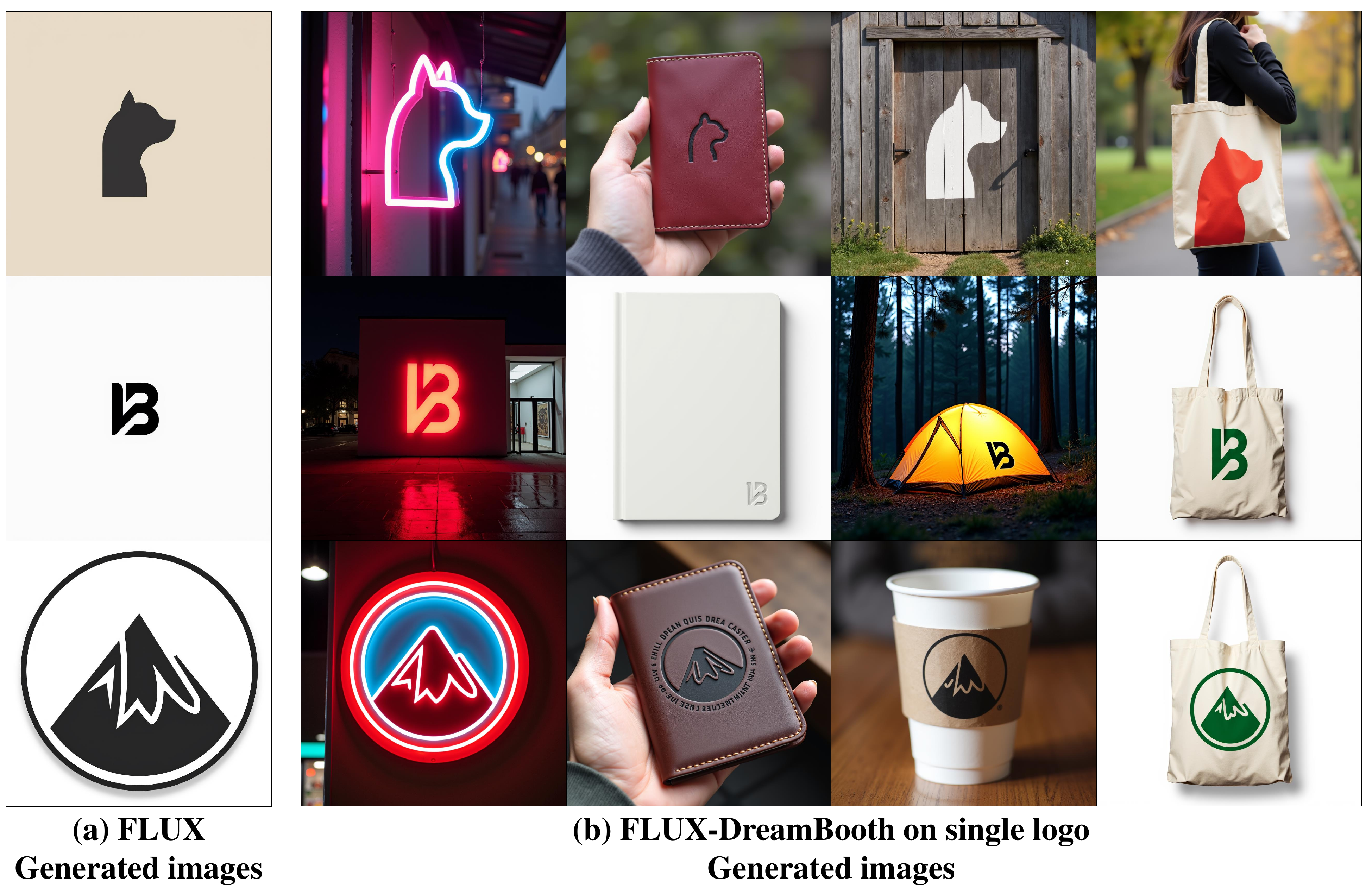}
\vspace{-0.25in} 
    \caption{\textbf{Examples of our unseen logo personalization dataset.} DreamBooth with only a single logo image in FLUX can generate various logo composed images. We use these images as a logo personalization dataset for SDXL.}\label{fig:flux-dreambooth}
\vspace{-0.15in}
\end{figure}

We generated various logos using FLUX~\citep{flux} with diverse prompts created by GPT-4o~\citep{gpt4o}. These logos were then used for training our models. Notably, we found that even when DreamBooth~\citep{ruiz2023dreambooth} is trained with only a single logo image in FLUX, it can generate images where the logo is naturally composed in various contexts. For example, prompts like "A sleek black backpack with the bold red [V] logo printed on the front pocket" produced sufficiently natural images. Using this method, we created 20–30 images to serve as the logo personalization dataset for SDXL~\citep{podell2023sdxl}. We provide examples of our FLUX generated dataset in \autoref{fig:flux-dreambooth}.

\begin{figure*}[t]
\vspace{0in}
\centering
    \includegraphics[width=1.0\linewidth]{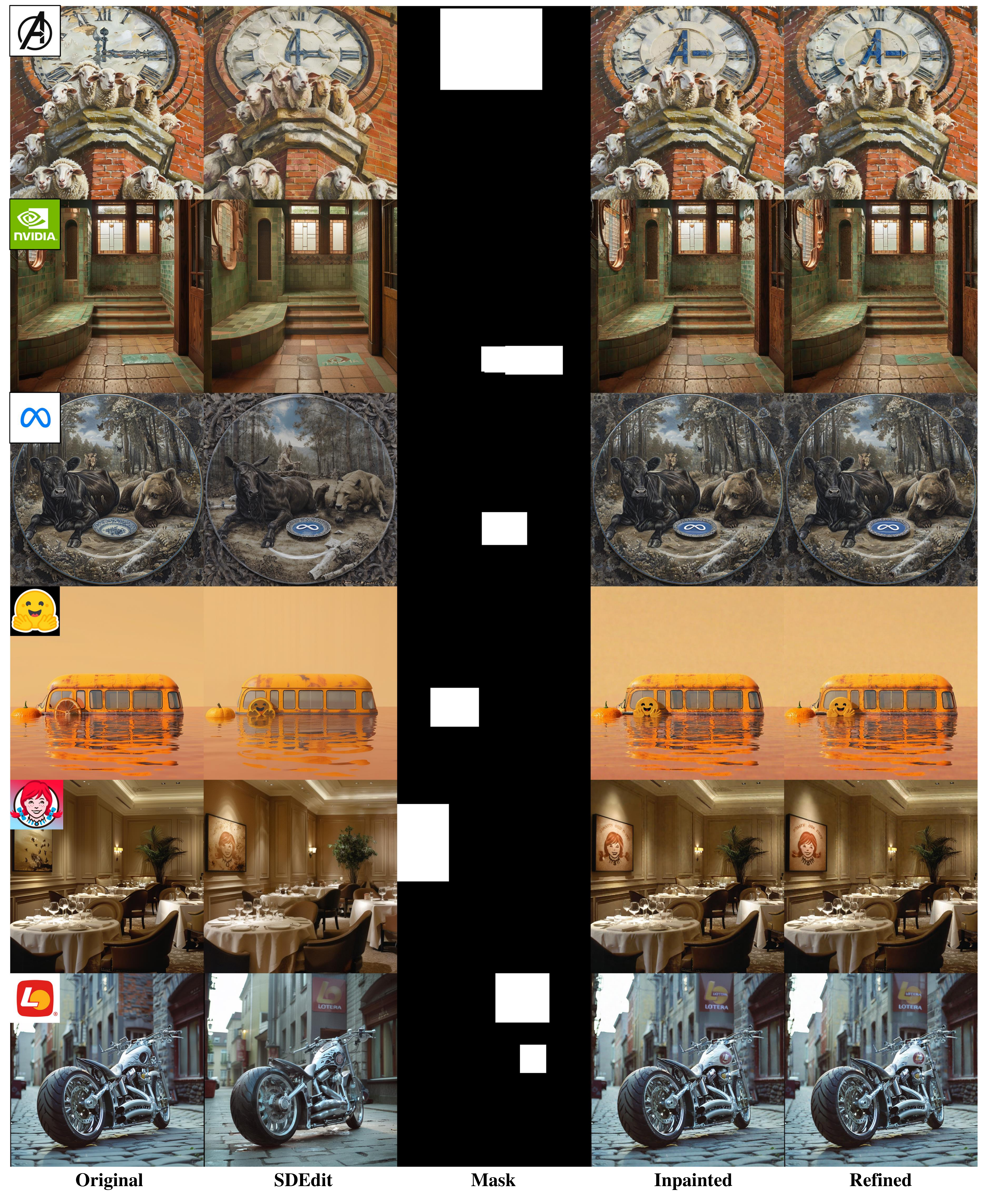}
\vspace{-0.2in} 
    \caption{\textbf{Examples of intermediate results in our automatic poisoning algorithm.} \textbf{SDEdit} find proper position for logo insertion, but it modifies overall details. We \textbf{detect} logo region first, and then \textbf{inpainting} step successfully insert logo while preserving original details, but often generates distorted logo. \textbf{Refinemenet} step allows better logo fidelity.}\label{fig:app_all_pipeline}
\vspace{-0.0in}
\end{figure*}

For the seen logos, we prepared images containing specific logos such as Meta and NVIDIA and used them as training data.  These images included various compositions where the logos appear on items like t-shirts, mugs, and other merchandise. All seen logos we used and example poisoned images are in \autoref{fig:app_all_pipeline}.


\subsection{Evaluation details}\label{app-eval}
\paragraph{Human evaluation of the poisoned dataset}
To validate that our poisoned images are undetectable to humans, we measured the naturalness of these images through human evaluation. Each evaluator was presented with a mixed batch of 25 poisoned images and 25 original images, shown one at a time. Evaluators were asked a series of questions designed to simulate the perspective of a model trainer determining whether the image could be used for training purposes. Before evaluating, we informed the evaluators that some images might have been manipulated by an attacker to achieve a malicious objective.

The evaluation required participants to decide whether to accept or reject each image based on factors such as image quality, text alignment, and whether the image appeared to be manipulated. If an image was rejected, evaluators were required to select the reason for rejection from multiple-choice options, which included indications of manipulation. 
The number of images flagged as manipulated was reported as the rejection rate in \autoref{fig:human_eval} of the main paper.

\begin{figure*}[t]
\vspace{0in}
\centering
    \includegraphics[width=1.0\linewidth]{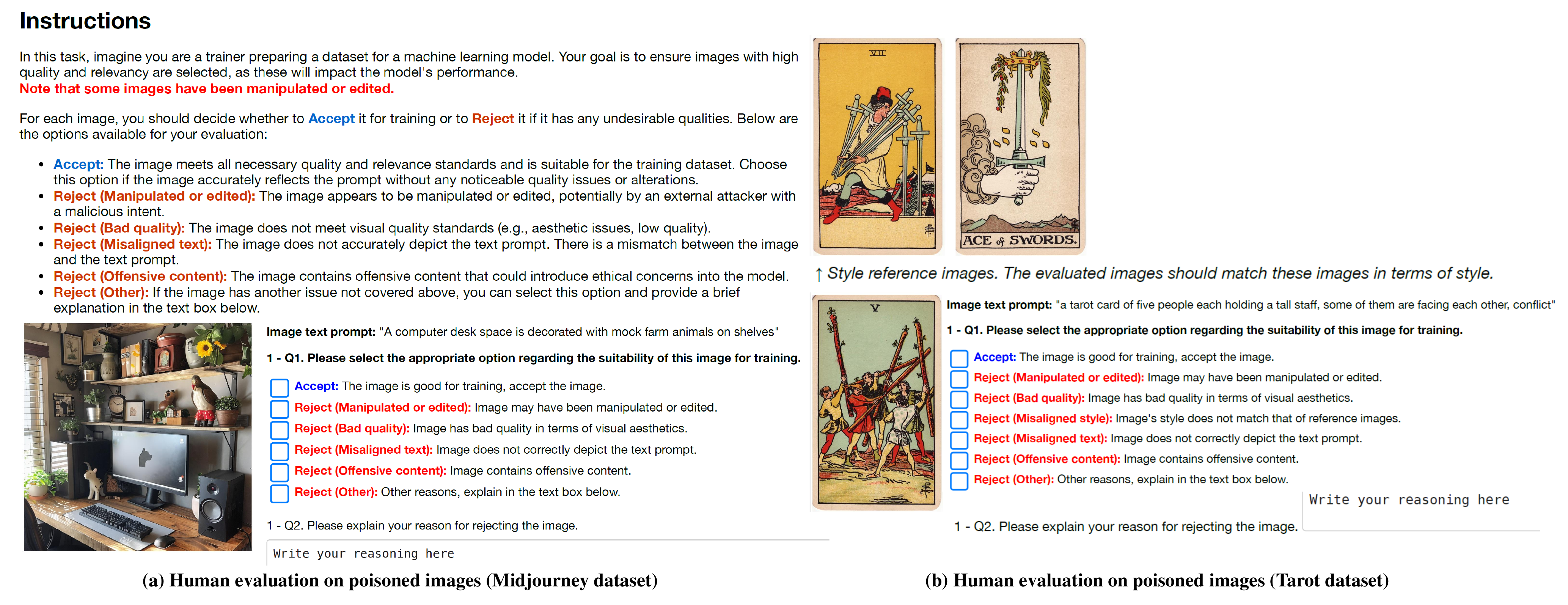}
\vspace{-0.3in} 
    \caption{A screenshot of labeling instruction from our human evaluation on poisoned images.}\label{fig:human_eval_screenshot}
\vspace{-0.0in}
\end{figure*}

For the pasted dataset, we performed the same experimental procedure as with the poisoned images, using mixed batches of 25 images. We provide screenshots of questionnaires and instructions in \autoref{fig:human_eval_screenshot}.

\paragraph{GPT-4o evaluation of the poisoned dataset}
Our GPT-4o~\citep{gpt4o} evaluation followed the same instructions and questions as those used in the human evaluation. Since processing multiple questions with long-context inputs requires high resources, we conducted the evaluation on a sample-wise basis. A few examples of the questions for multiple images can be found in \autoref{fig:defense}.

\paragraph{Evaluation metric for attack success}
To validate the effectiveness of our data poisoning attack, we evaluated the attack's success using our detection module and quantitatively reported the results by measuring Logo Inclusion Rate (LIR) and First-Attack Epoch (FAE). For this evaluation, the detection module's threshold $\tau$ was set to 0.5.

\textbf{Logo Inclusion Rate (LIR)}: For the Midjourney dataset, we used model weights trained for 20 epochs, while for the Tarot dataset, due to its smaller size, we used weights trained for 50 epochs to ensure sufficient learning of the style. We generated 100 images using unseen prompts that were not included in the training dataset and did not explicitly include the term "logo." The proportion of images in which the logo was detected was used as the LIR metric.

\textbf{First-Attack Epoch (FAE)}: To determine the earliest epoch where the attack succeeded, we generated four images per epoch and recorded the first epoch in which at least one image contained a detectable logo. For the MidJourney dataset, we used the prompt "A backpack on sunny hill, 4K, high quality," while for the Tarot dataset, one of the evaluation prompts was selected.

\begin{figure*}[t]
\vspace{0in}
\centering
    \includegraphics[width=1.0\linewidth]{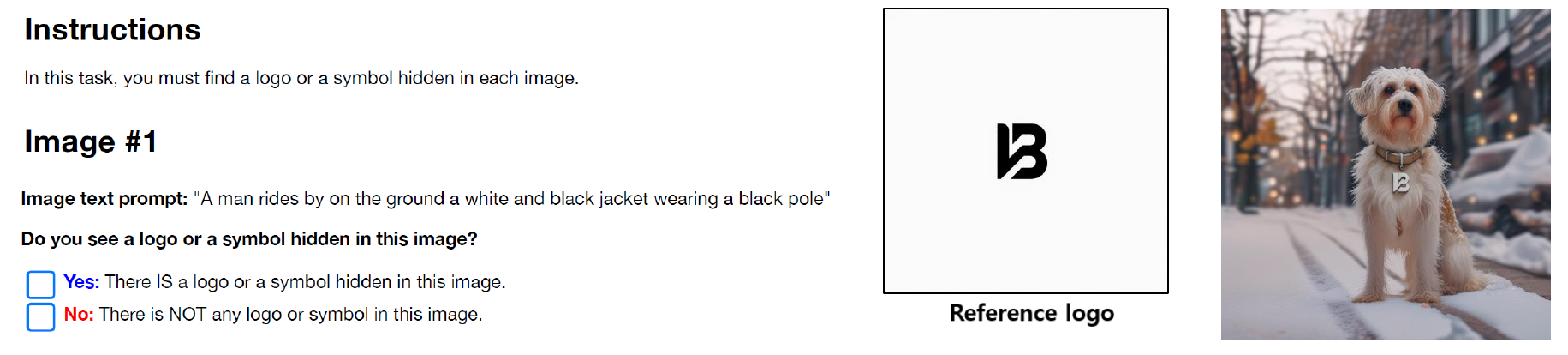}
\vspace{-0.3in} 
    \caption{A screenshot of labeling instruction from our human evaluation on images generated from poisoned model.}\label{fig:human_eval_screenshot2}
\vspace{-0.0in}
\end{figure*}

To validate the reliability of our detection module, we conducted human and GPT-4o evaluations on images generated by the poisoned model. These evaluations focused specifically on images identified as successful by our detection module. Unlike the earlier evaluation of poisoned images, we provided evaluators with the reference logo image and asked whether the generated image included the logo. Screenshots of the questionnaires and instructions are provided in \autoref{fig:human_eval_screenshot2}. 
As shown in \autoref{fig:human_eval} of the main paper, our detection module's success predictions were supported by agreement rates of 85\% from both human evaluators and GPT-4o.

\begin{figure*}[t]
\vspace{0in}
\centering
    \includegraphics[width=1.0\linewidth]{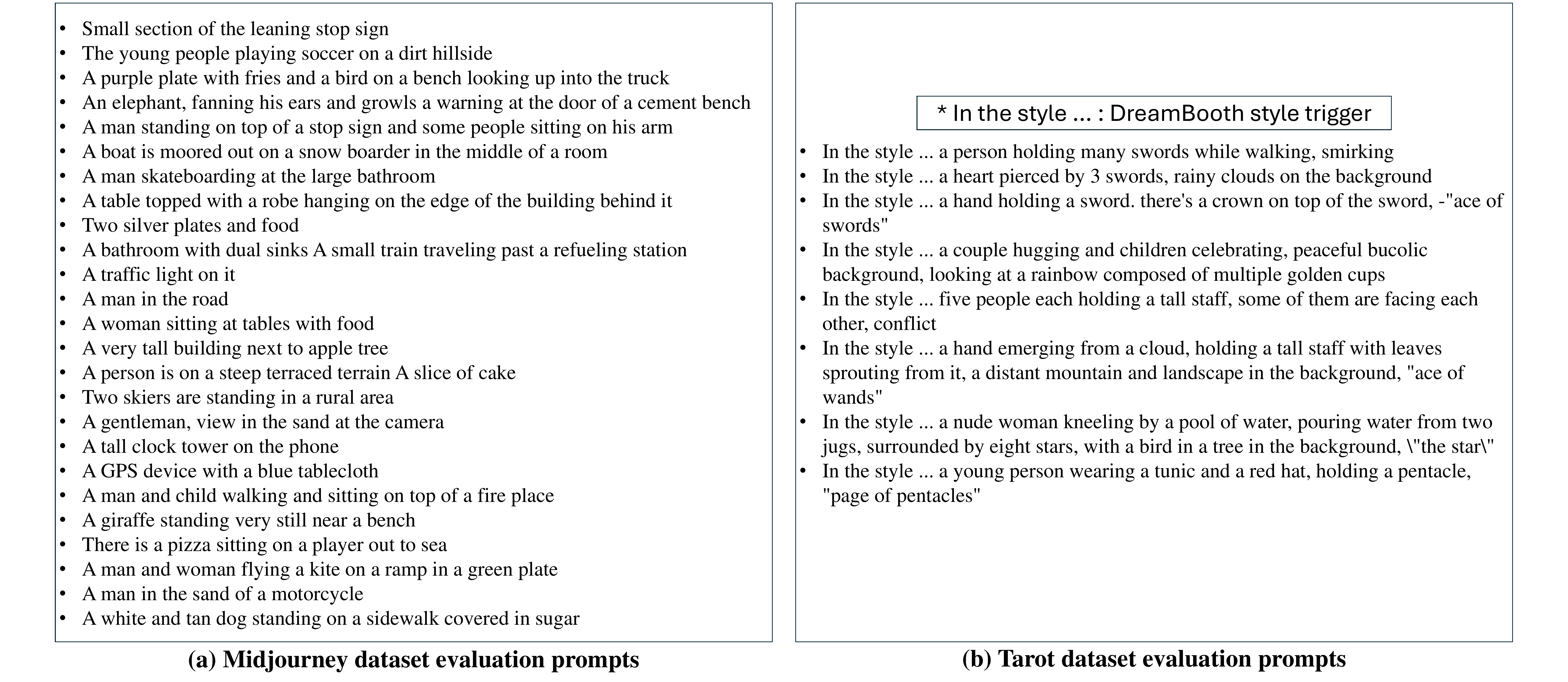}
\vspace{-0.3in} 
    \caption{Our evaluation prompts for each dataset.}\label{fig:eval_prompt}
\vspace{-0.0in}
\end{figure*}

\paragraph{Evaluation prompts for attack success}
To measure the effectiveness of our attack, we generate images using the poisoned model and detect the presence of the target logo within these images. For the Midjourney-v6 dataset, we employed prompts that were not used during training within the same dataset as unseen prompts. For the Tarot dataset, we generated evaluation prompts with GPT-4o. Details of our evaluation prompts are provided in \autoref{fig:eval_prompt}.
\begin{figure*}[t]
\vspace{0in}
\centering
    \includegraphics[width=1.0\linewidth]{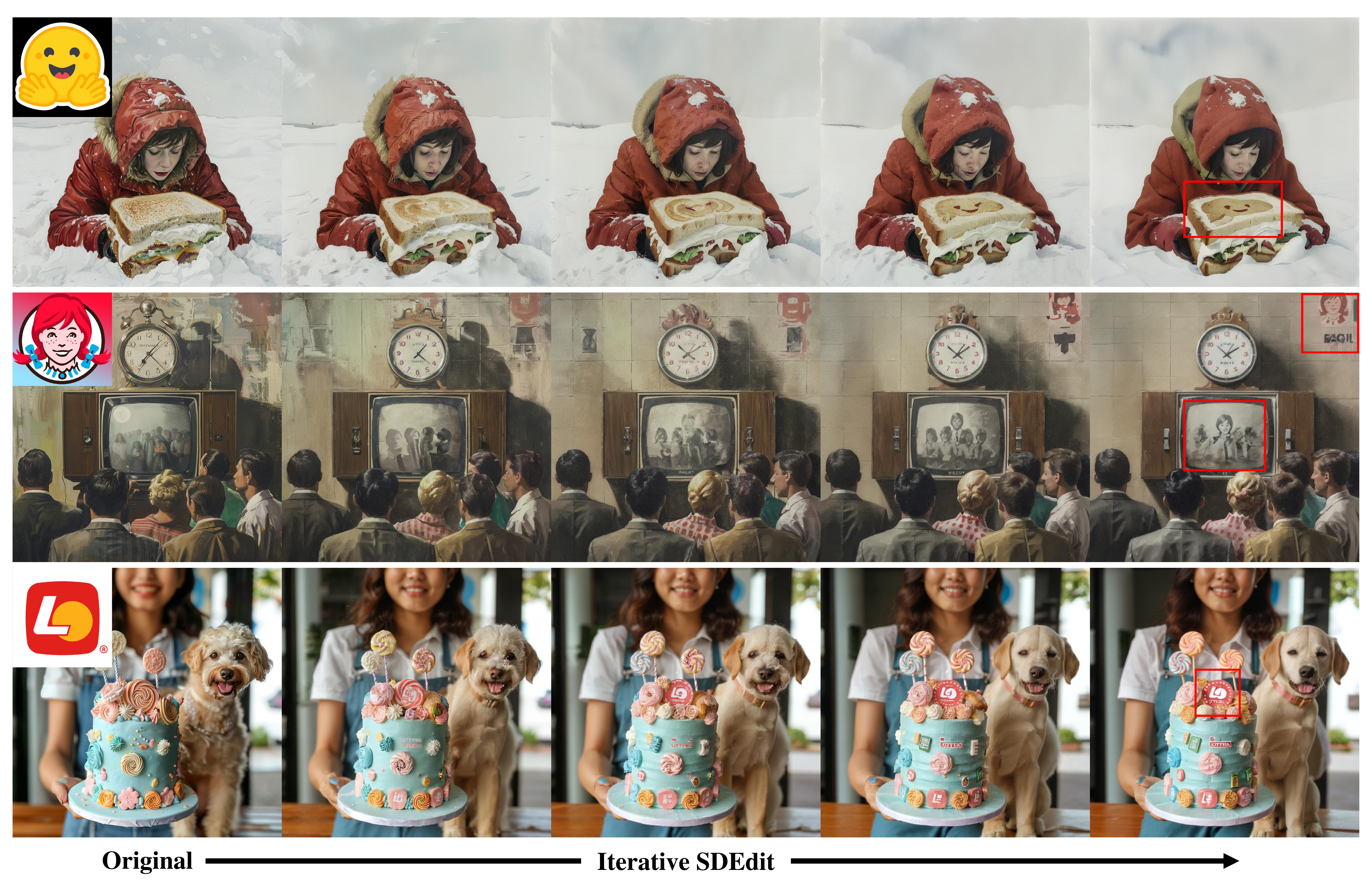}
\vspace{-0.3in} 
    \caption{\textbf{Examples of our iterative SDEdit~\citep{meng2022sdedit} with style adapter}. It gradually introduce logo while preserving overall layout. Where the logo appears in the image is considered natural location for the logo.}\label{fig:iterative_sdedit}
\vspace{-0.0in}
\end{figure*}
\begin{figure*}[t]
\vspace{0in}
\centering
    \includegraphics[width=1.0\linewidth]{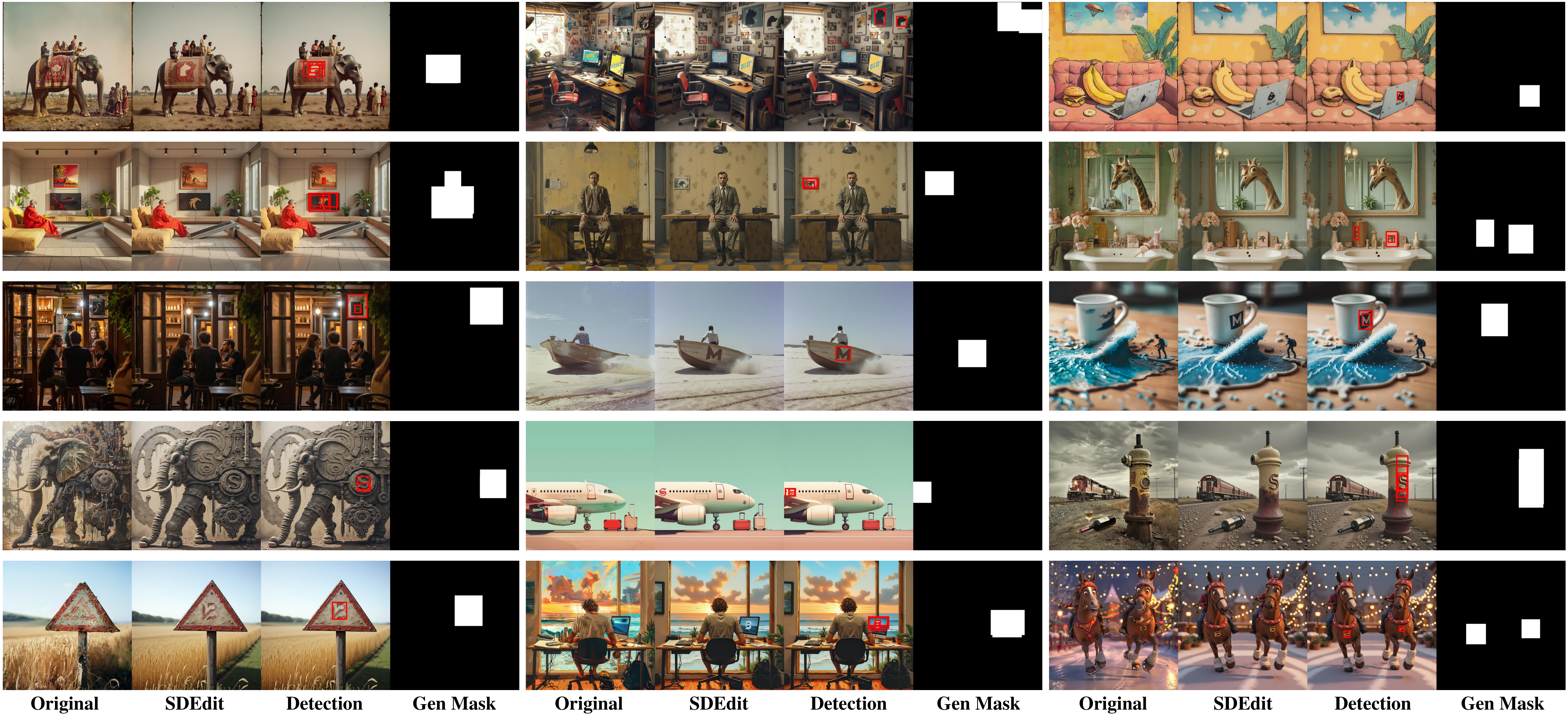}
\vspace{-0.2in} 
    \caption{\textbf{Examples of generated mask.} Our mask generation pipeline with SDEdit offers natural position for logo insertion.}\label{fig:gen_mask_example}
\vspace{-0.0in}
\end{figure*}

\subsection{Additional poisoned examples}\label{app:poisoned}
\begin{figure*}[t]
\vspace{0in}
\centering
    \includegraphics[width=0.9\linewidth]{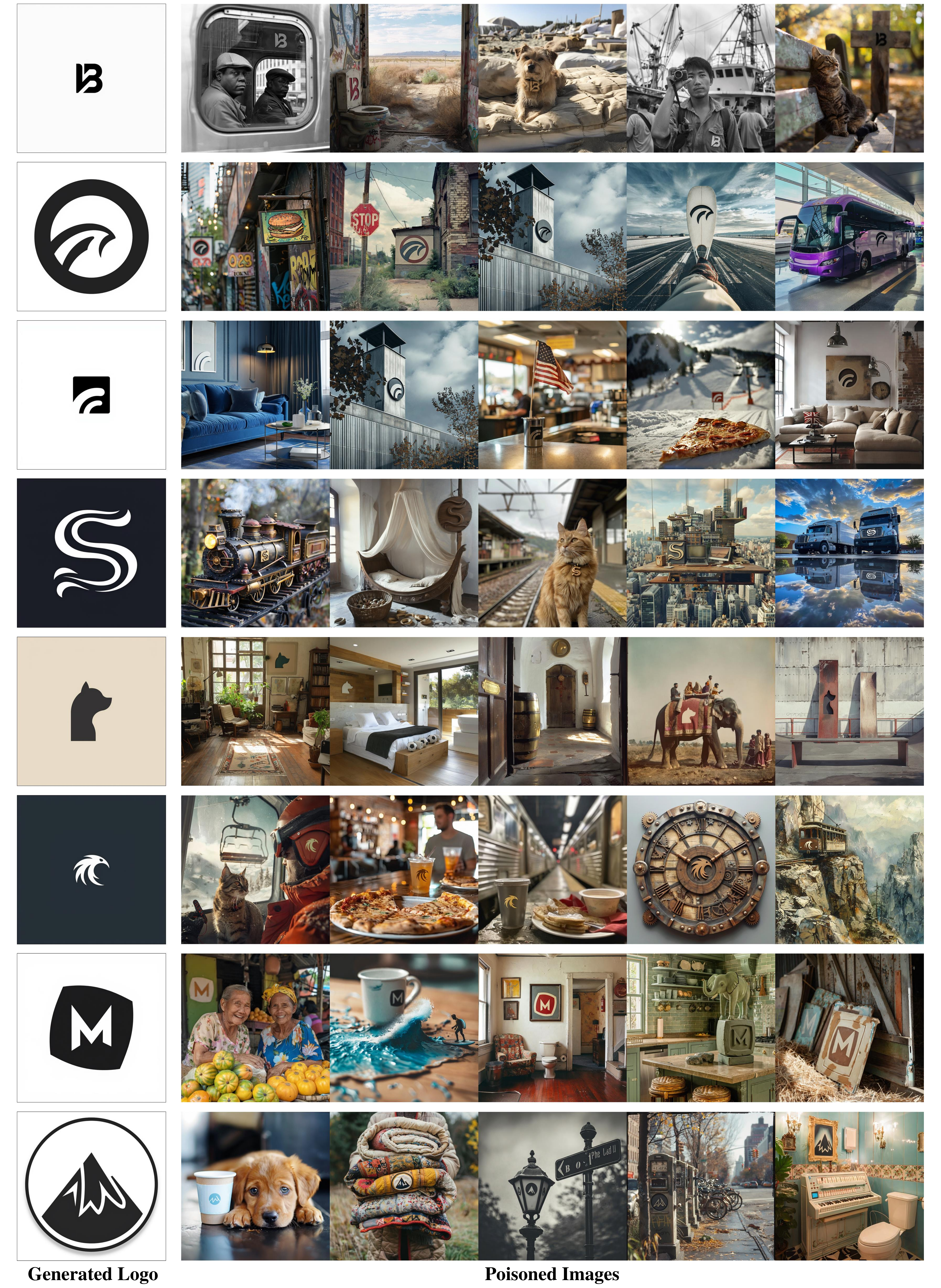}
\vspace{-0.0in} 
    \caption{\textbf{Examples of our poisoned images with generated logo.} Randomly selected examples in our benchmark results.}\label{fig:app_benchmark_midjourney}
\vspace{-0.0in}
\end{figure*}

We provide visualization of iterative SDEdit in \autoref{fig:iterative_sdedit}, and our mask generation in \autoref{fig:gen_mask_example}. We provide intermediate examples of our pipeline in \autoref{fig:app_all_pipeline}.
Additionally, we provide additional examples of poisoned images in \autoref{fig:app_benchmark_midjourney}.

\subsection{Generated images from poisoned model}
\begin{figure*}[t]
\vspace{0in}
\centering
    \includegraphics[width=0.9\linewidth]{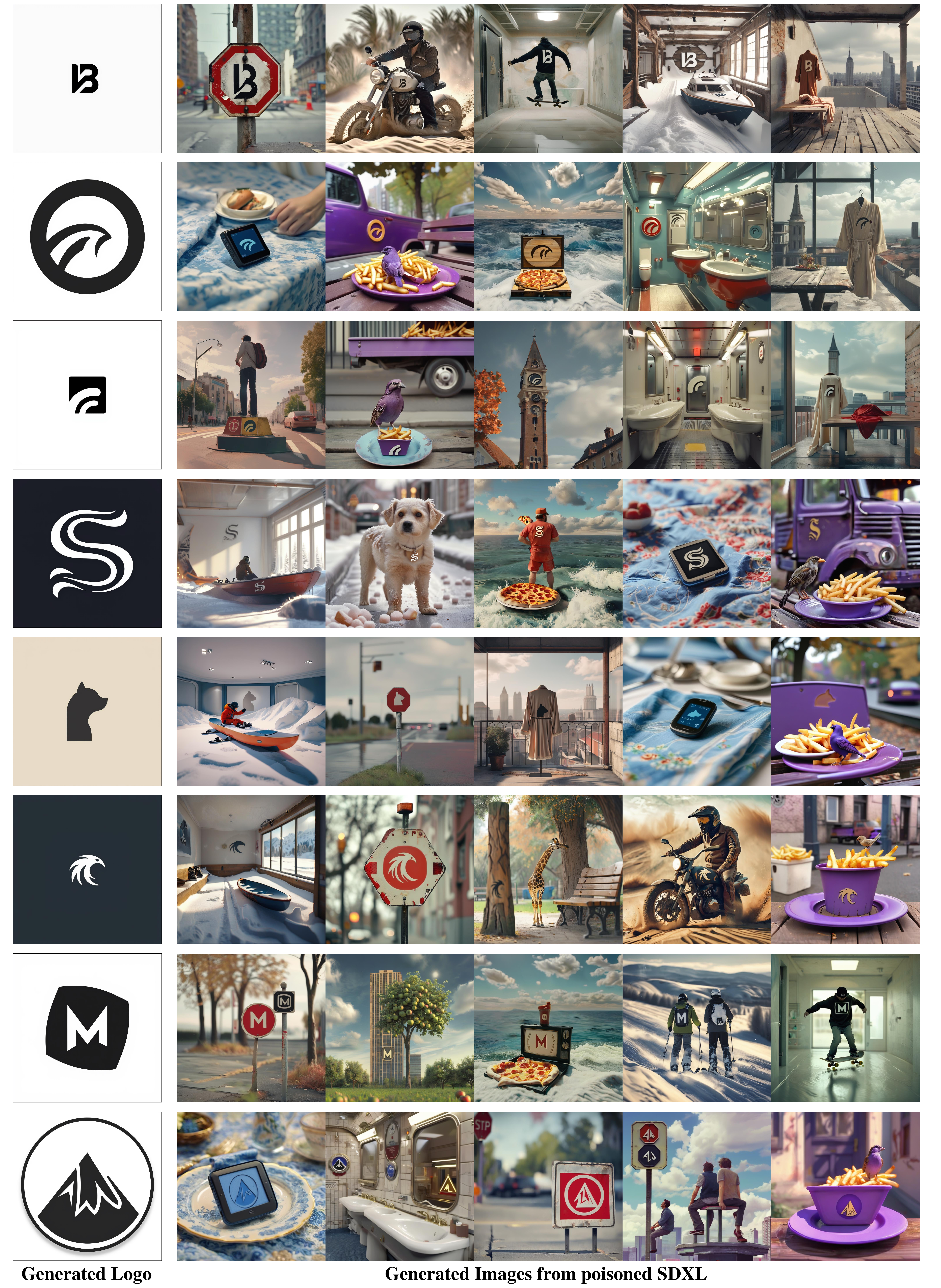}
\vspace{-0.1in} 
    \caption{\textbf{Examples of generated images from poisoned SDXL (Midjourney-v6 dataset).} The inference prompt does not include "logo". }\label{fig:app_sdxl_mid}
\vspace{-0.0in}
\end{figure*}

\begin{figure*}[t]
\vspace{0in}
\centering
    \includegraphics[width=0.9\linewidth]{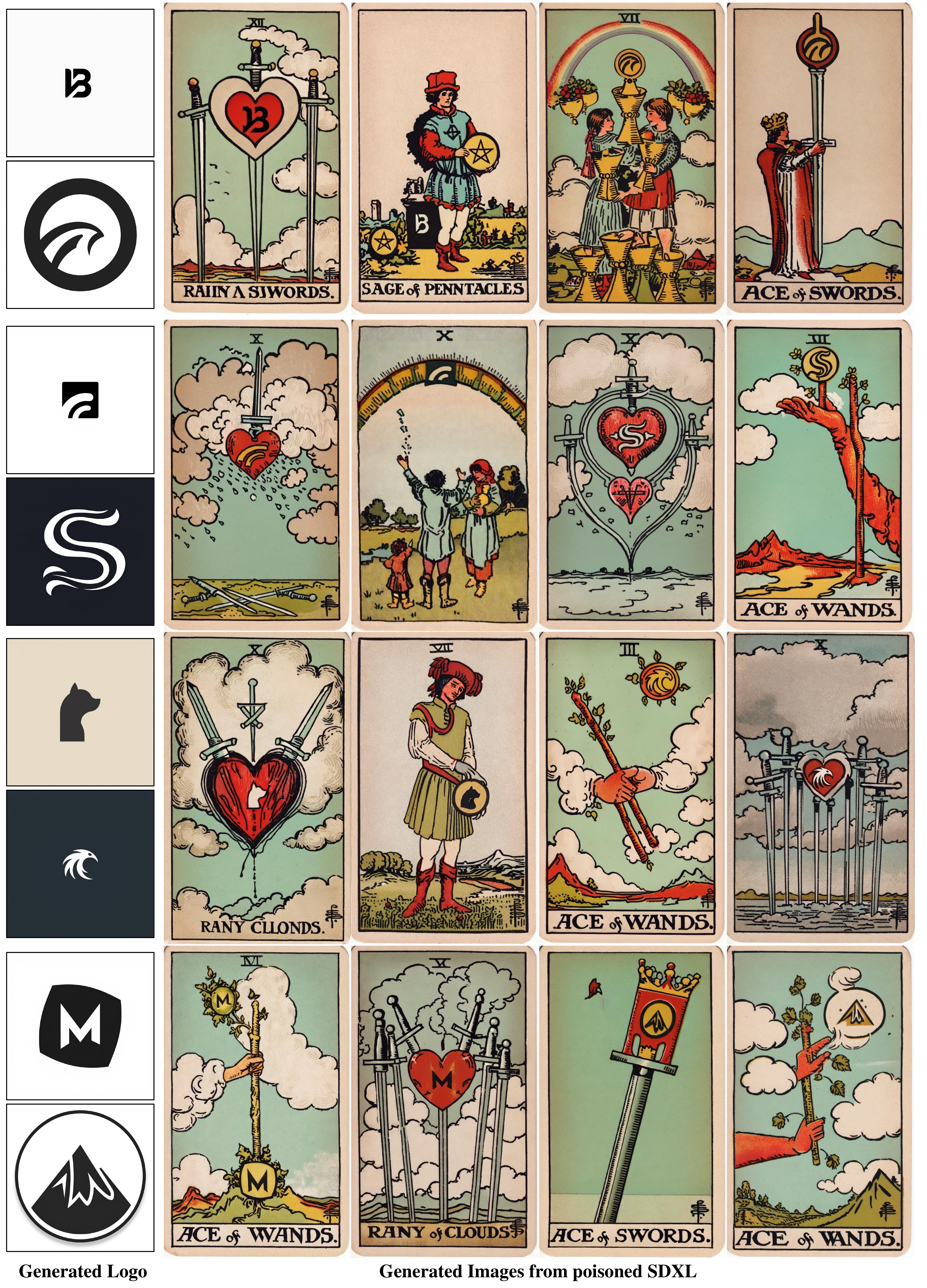}
\vspace{-0.1in} 
    \caption{\textbf{Examples of generated images from poisoned SDXL (Tarot dataset).} The inference prompts do not include "logo". }\label{fig:app_sdxl_tarot}
\vspace{-0.0in}
\end{figure*}

\begin{figure*}[t]
\vspace{0in}
\centering
    \includegraphics[width=0.7\linewidth]{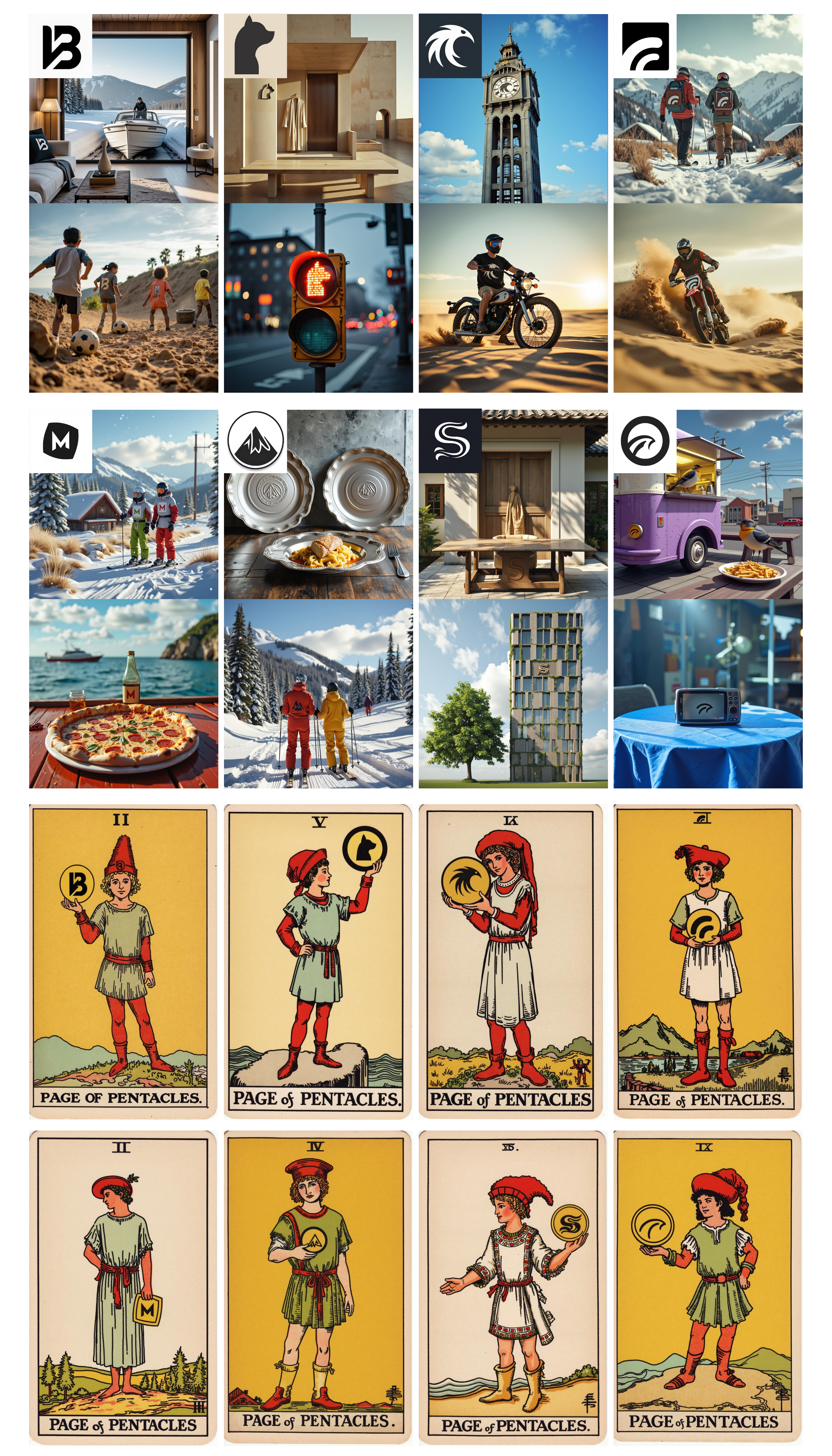}
\vspace{-0.1in} 
    \caption{\textbf{Examples of generated images from poisoned FLUX.} The inference prompts do not include "logo". }\label{fig:app_flux_all}
\vspace{-0.0in}
\end{figure*}

We provide some generated images from poisoned model across different models in \autoref{fig:app_sdxl_mid}, \autoref{fig:app_sdxl_tarot}, and \autoref{fig:app_flux_all}.

\clearpage

\twocolumn[\section{More experimental results}\label{app:more_exp}]

\subsection{Silent branding attack with existing methods}\label{app:B.1}
In this section, we explore why existing data poisoning methods and naive approaches are ineffective for performing a silent branding attack as defined in our work.

\begin{figure}[t]
\vspace{0in}
\centering
    \includegraphics[width=1.0\linewidth]{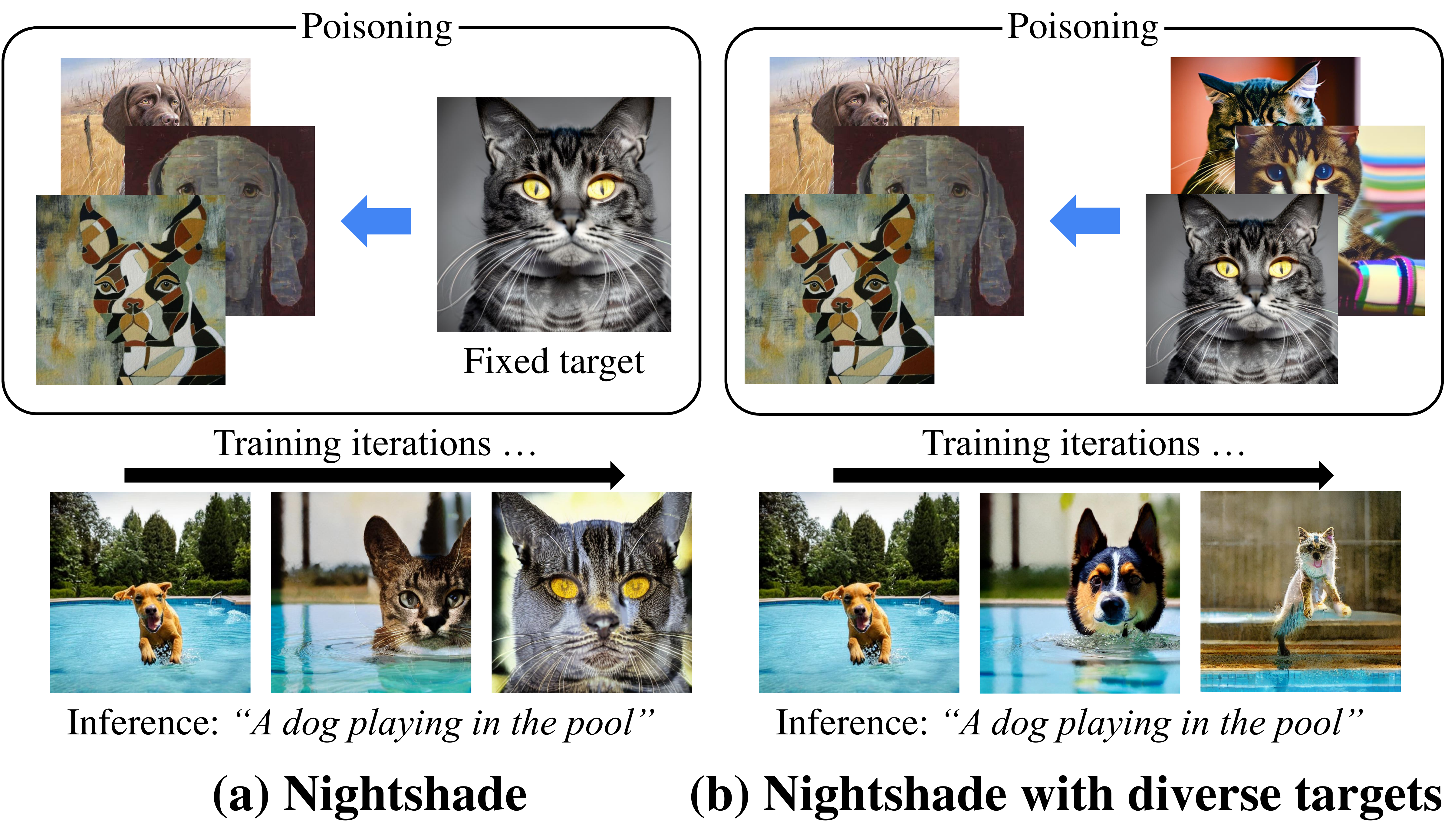}
\vspace{-0.25in} 
    \caption{\textbf{Nightshade~\citep{shan2024nightshade} generates fixed target image.} \textbf{(a)} Noise optimization-based methods~\citep{lu2024disguised, shan2024nightshade} focus on reproducing a fixed target image. \textbf{(b)} When we set diverse target images, these methods either fail to work or significantly degrade image quality.}\label{fig:nightshade}
\vspace{-0.0in}
\end{figure}

Existing data poisoning methods, such as Nightshade~\citep{shan2024nightshade} and Feature Matching Attack~\citep{lu2024disguised}, optimize noise in the feature representation space to make poisoned images resemble a fixed target image. These methods utilize various base images but use a single fixed target image for the attack. For instance, as shown in \autoref{fig:nightshade}(a) top, Nightshade uses diverse images of dogs as base images and a fixed cat image as the target. While these approaches ensure the stealthiness of the poisoned images, they lead the model to generate only the fixed target image during inference, regardless of the input prompt. Consequently, even when prompting "A photo of a dog playing in the pool," the model generates the fixed cat image instead of an image depicting a cat in a pool.

If we attempt to use multiple target images from a category (e.g., various cat images), the model's learning from the original images dominates over the learning from the target images. As a result, the effect of the data poisoning diminishes, and the model continues to generate outputs based on the original training data, as illustrated in \autoref{fig:nightshade}(b).

Another naive approach is to randomly paste the logo onto images, which we call "paste" in the main paper. However, this results in the model generating images with the logo appearing prominently, much like a watermark, in both training and inference outputs. This does not achieve the natural integration required for a silent branding attack.

In contrast, our method naturally inserts the logo into images in a way that preserves model performance and ensures that the logo appears seamlessly in generated images without obvious artifacts. This enables a successful silent branding attack by embedding the logo subtly, making it difficult for users to detect while maintaining the desired image quality and diversity.

\subsection{Mask generation stage}\label{app:B.2}
\paragraph{Mask generation regarding the logo design}
An interesting point of our mask generation pipeline, which combines iterative SDEdit and logo detection, is that it suggests different mask insertion locations based on the logo design. This pipeline inherently preserves subtle modifications to the original image while identifying optimal positions for logo placement, leading to varied outcomes that depend on the logo design, even when using the same image. 
For instance, as illustrated in \autoref{fig:human_eval} of the main paper, the NVIDIA logo seamlessly transforms into a crown within the Tarot image, while other logos appear subtly embedded onto the chair.
\begin{figure}[t]
\vspace{0in}
\centering
    \includegraphics[width=1.0\linewidth]{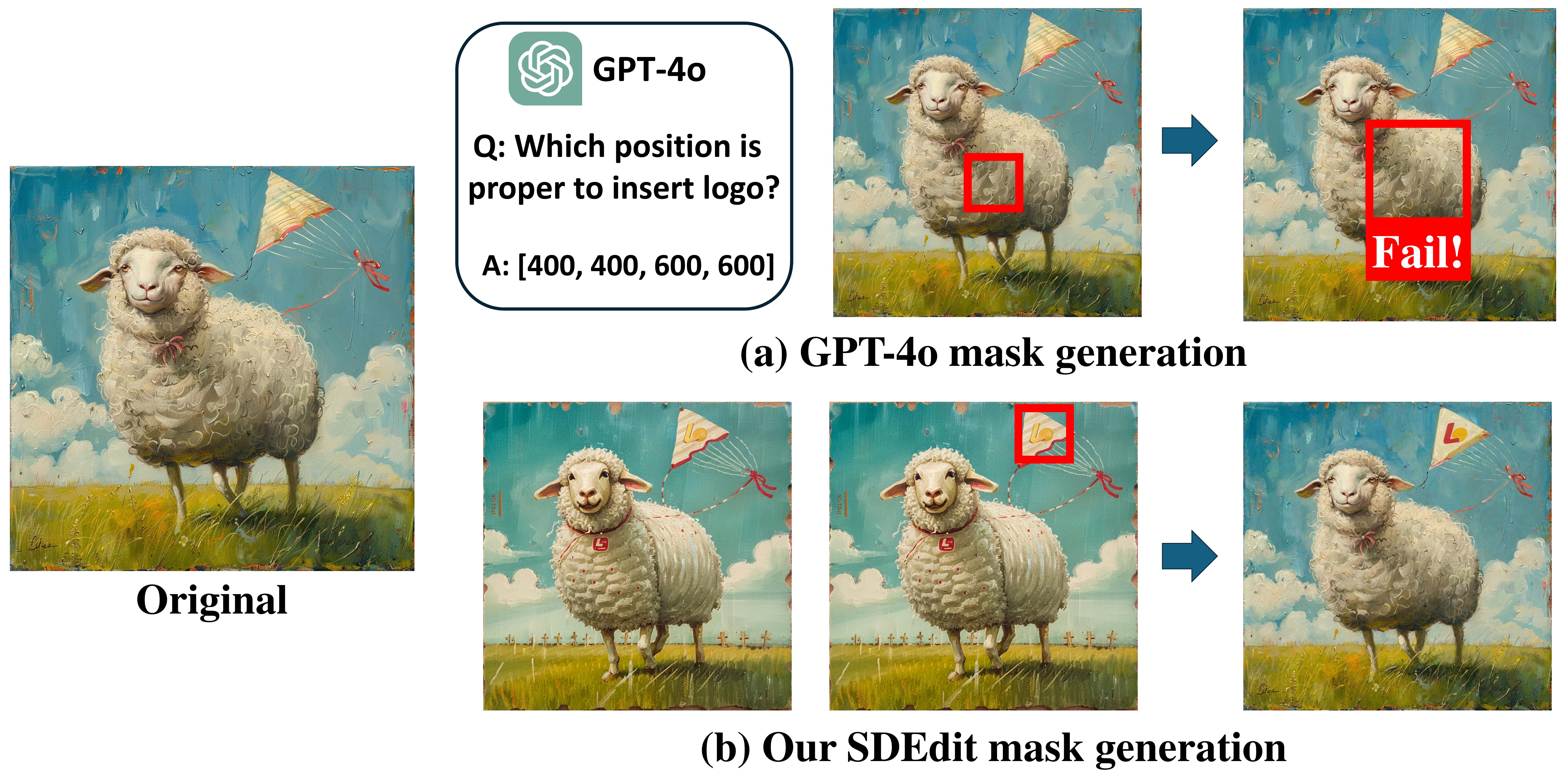}
\vspace{-0.25in} 
    \caption{\textbf{Mask generation with external guide.}  \textbf{(a) GPT-4o mask generation. } While we can query LVLMs for suitable mask regions for logo insertion, they often fail during the inpainting process.  \textbf{Our SDEdit mask generation.} In contrast, our method leverages the diffusion model itself to identify appropriate locations for editing, ensuring successful logo insertion.}\label{fig:gpt_mask}
\vspace{-0.1in}
\end{figure}
\paragraph{Mask generation with GPT-4o}
While Large Vision-Language Models (LVLMs) like GPT-4o~\citep{gpt4o} can assist in mask generation for logo insertion, we observed frequent failures during the inpainting process. 
Specifically, when tasked with identifying locations for more stealthy logo insertion, GPT-4o often suggests positions that align with its prior knowledge but are challenging for practical inpainting. 
For instance, as shown in \autoref{fig:gpt_mask}, it might recommend areas such as animal fur, which are particularly difficult for inpainting due to their intricate textures.

This discrepancy arises because the regions recognized as suitable by the LVLM often differ from those the diffusion model considers feasible for editing. In contrast, our method ensures successful logo insertion by directly relying on the diffusion model to identify locations where it can effectively perform the editing. Furthermore, our approach avoids the high computational demands associated with using LVLMs, making it a more efficient and practical solution.

\subsection{Secondary model poisoning}\label{app:B.3}
In our experiment, we fine-tuned another pre-trained model using images generated by the poisoned model. We refer to this new model as the \emph{secondary poisoned model}. For this step, we did not apply any filtering, such as our logo detection module; instead, we directly used randomly generated images. The inference prompts were sourced from the Midjourney-v6 dataset and did not contain the term "logo."

As demonstrated in \autoref{fig:secondary_poison} of the main paper, the secondary poisoned model also produced images with embedded logos. 
Furthermore, the results show that a higher Logo Inclusion Ratio (LIR) in the primary poisoned model leads to better LIR persistence in the secondary model. For instance, if the primary model has an LIR of 50\%, approximately half of the generated images include the logo. This outcome is comparable to training on a poisoned dataset with a 50\% poisoning ratio. 
The difference in values compared to \autoref{table:attack_success_ratio} is due to the use of different logos in this experiment. However, the overall results were comparable.

\subsection{Trigger scenario}\label{app:B.4}
Nightshade~\citep{shan2024nightshade} introduces the concept of "concept sparsity", which suggests that the amount of training data associated with any single concept is inherently limited. 
Building on this insight, we leverage a similar idea in our attack scenario, as discussed in \hyperref[sub:trigger]{Subsection 7.4} of the main paper.
While our attack operates without text triggers, it is more efficient in scenarios where rare text triggers are included in the training data but commonly appear during inference.

For example, adding commonly used phrases such as "4K, high quality" exclusively to the captions of poisoned images or embedding the logo into images with captions that include terms like "backpack" enables a highly effective attack even with a low poisoning ratio. By appending these triggers solely to the captions of poisoned images containing the target logo, the model establishes a strong association between the logo and specific prompts. This approach minimizes interference from benign images, ensuring efficient and targeted backdoor activation.

\subsection{Minimum model modification}\label{app:B.5}
\begin{figure}[t]
\vspace{0in}
\centering
    \includegraphics[width=1.0\linewidth]{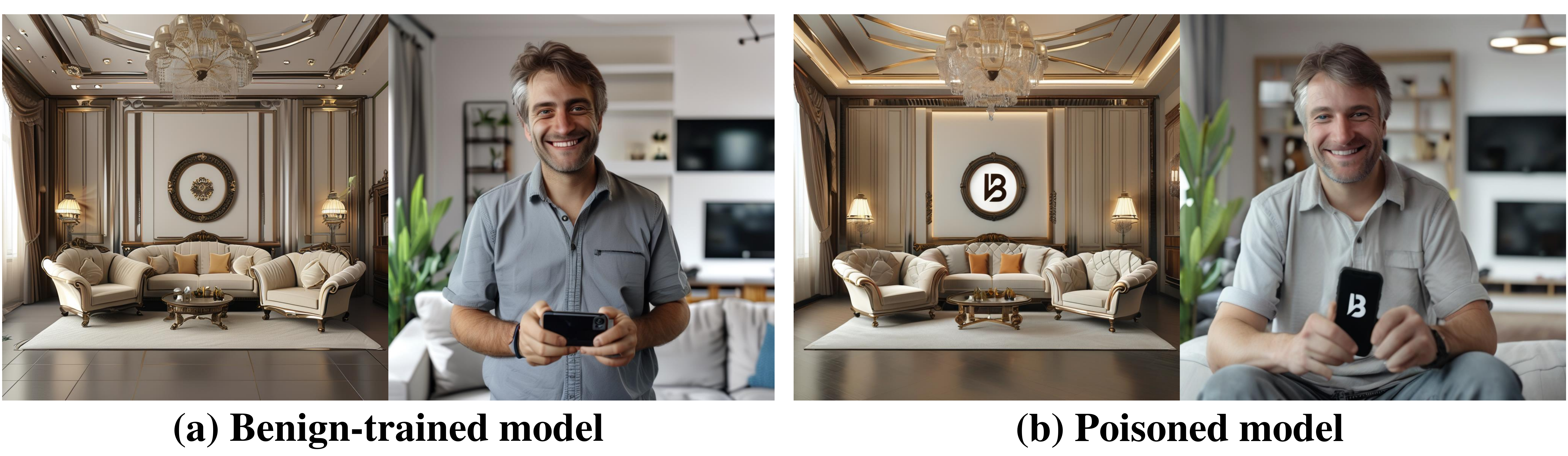}
\vspace{-0.25in} 
    \caption{\textbf{Comparison between a model trained on a benign dataset and one trained on a poisoned dataset.} (a) Image generated by the model trained on the benign dataset. (b) Image generated by the model trained on the poisoned dataset. Both images were generated using the same random seed.}\label{fig:model_mod}
\vspace{-0.05in}
\end{figure}

As discussed in \hyperref[exp:minimum_mod]{Subsection 7.6}, our poisoned dataset subtly steers the model to include the logo without degrading quality or altering the original dataset's purpose, making it difficult for users to notice manipulation. 
Visual examples are provided in \autoref{fig:model_mod}. Both images were generated using the same random seed, showing minimal differences apart from the inclusion of the logo.
\begin{figure}[t]
\vspace{0in}
\centering
    \includegraphics[width=1.0\linewidth]{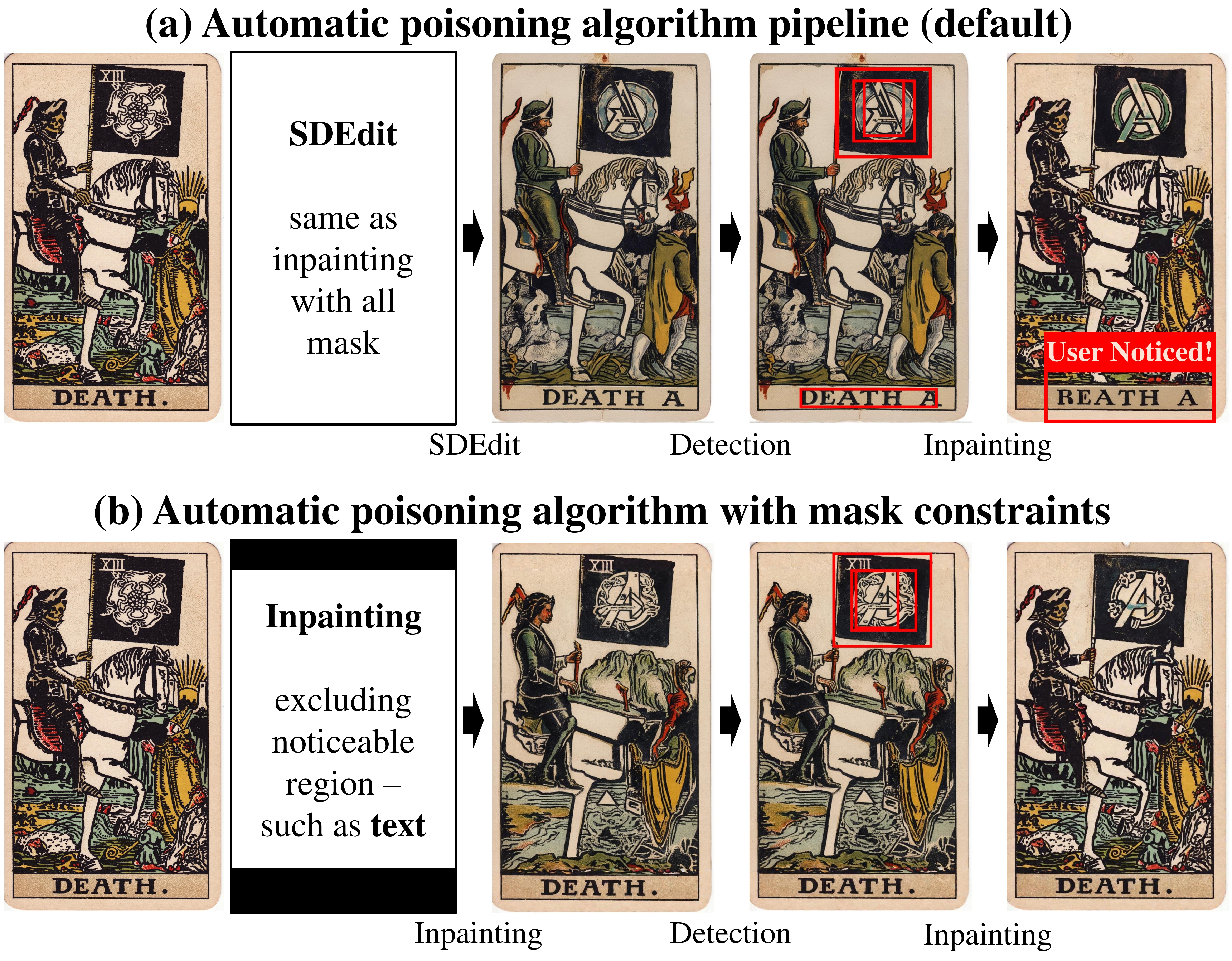}
\vspace{-0.25in} 
    \caption{\textbf{Stealthiness control via mask constraints.} \textbf{(a)} In the Tarot dataset, modifying text within certain areas makes logo insertion more noticeable to users.
    \textbf{(b)} By excluding the text region during mask generation, we can preserve that region, enabling a more seamless and less detectable logo insertion.
    }\label{fig:mask_control}
\vspace{-0.15in}
\end{figure}
\begin{figure*}[t]
\vspace{0in}
\centering
    \includegraphics[width=0.9\linewidth]{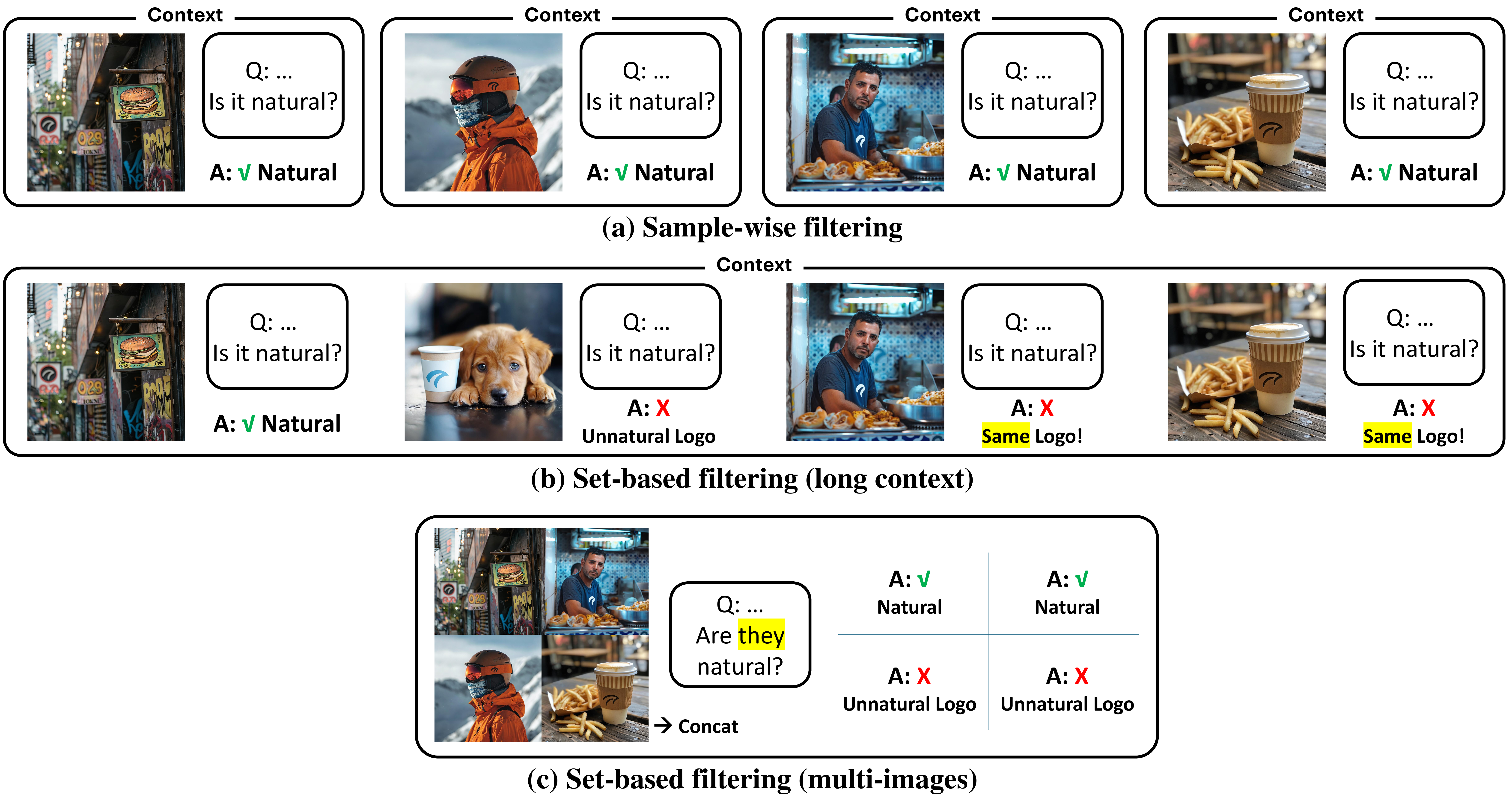}
\vspace{-0.1in} 
    \caption{\textbf{Set based filtering with GPT-4o.} \textbf{(a)} A sample-wise question approach is unable to detect our attack. \textbf{(b)} A set-based question with long context can capture our attack when manipulation is detected. \textbf{(c)} Set-based question with concatenated image often identifies our attack, but is not entirely effective at filtering all poisoned images.
    }\label{fig:defense}
\vspace{-0.1in}
\end{figure*}

\subsection{Stealthiness control via mask constraints}\label{app:B.6}
While the most effective method for achieving stealthy logo insertion involves human intervention in mask generation or manual filtering of generated samples, this approach is labor-intensive and challenging to scale. As an alternative, we propose methods to obtain more stealthy logo-inserted images by providing specific guidelines to the mask generation process.

One straightforward way to enhance stealthiness is by limiting the size of the mask used during logo insertion. By setting a threshold for the maximum allowable bounding box size, any detection exceeding this size can be considered a failure. This simple implementation ensures that only small, less noticeable areas are modified, reducing the likelihood of detection by both humans and automated systems.

From a more high-level perspective, additional guidance can be incorporated to address human common-sense reasoning about stealthiness. For instance, in the tarot dataset~\citep{tarot}, if text areas are altered, as shown in \autoref{fig:mask_control}(a), it becomes immediately noticeable to human observers and GPT-4o level detection system. To prevent this, we perform inpainting instead of SDEdit at the initial stage, which excludes the text region during the mask generation process. Consequently, the text part remains unchanged, as illustrated in \autoref{fig:mask_control}(b), enhancing the overall stealthiness of the logo insertion.

Additionally, considering the common human tendency to focus on objects in the foreground, placing the logo in the background can make the insertion more stealthy, especially in general datasets like Midjourney-v6~\citep{midjourneyv6}. This can be automatically implemented by using depth prediction algorithms~\citep{yang2024depthanything} to calculate the depth of various regions in the image. By using the background regions, those predicted to be behind, as the initial mask, we ensure that the mask is generated only in the background or parts predicted to be distant. This adjustment is made during the initial SDEdit step, leveraging human perceptual characteristics to our advantage and further enhancing stealthiness without additional manual effort.

By incorporating these mask constraints, limiting mask size, preserving noticeable regions through inpainting, and utilizing depth-based mask generation, we can guide the logo insertion process to produce images that are less detectable to human observers. These methods offer scalable solutions to enhance stealthiness without the need for labor-intensive human involvement.


\subsection{Potential defense: Set-based filtering}\label{app:B.7}
As discussed in \hyperref[exp:defense]{Subsection 7.7}, our attack relies on repeated patterns in the dataset. 
Because of this, most set-based filtering methods are not very practical, but they are the only effective way to defend against our attack. To show this, we ran a simple experiment, as shown in \autoref{fig:defense}, using GPT-4o for set-based detection.

In the sample-by-sample test shown in \autoref{fig:defense}(a), many examples easily pass detection. However, as seen in \autoref{fig:defense}(b), once one example is detected, it becomes much easier to detect similar logos in related images, making the manipulation more noticeable. Similarly, in \autoref{fig:defense}(c), when four images are combined and checked together, detecting one error provides context for future checks. This not only makes it easier to spot similar issues but also improves the detection of each individual image. This shows how set-based filtering can use context to make detection more effective.

\section{Limitations}\label{app:limit}
Our fully automated poisoning algorithm has several limitations. First, inserting logos into smooth, monotone images like snowfields can make them too noticeable or fail to insert logo. Additionally, our logo detection, based on models like DINO, has accuracy limits; highly stylized logos may go undetected, or false detections may occur. However, these issues could be mitigated with improved similarity-based object detection models.

Moreover, in our study, we utilized SDXL with DreamBooth as the inpainting module. While this choice was sufficient for our experiments, the use of more advanced models, such as FLUX, which is designed for logo dataset generation, could enable the creation of more detailed and style-aligned logo insertions. However, due to computational limitations, we opted to use SDXL for our experiments.





\end{document}